\documentclass{article} 
\usepackage{collas2026_conference,times}
\usepackage{easyReview}
 
\usepackage[utf8]{inputenc}
\usepackage{microtype}
\usepackage{graphicx}
\usepackage{subcaption}
\usepackage{booktabs}
\usepackage{subscript}
\usepackage{amssymb}
\usepackage{amsmath}
\usepackage{mathtools}
\usepackage{amsthm}
\usepackage{tabularx}
\usepackage{url}
\usepackage{float}
\usepackage{placeins}
\usepackage{wrapfig}
\newcommand{\rot}[1]{\rotatebox{90}{{#1}}} 
\newcounter{algocounter}
\renewcommand{\thealgocounter}{\arabic{algocounter}}
\newenvironment{algorithmbox}[1]{%
  \refstepcounter{algocounter}%
  \bigskip\noindent
  \hrule\vspace{4pt}
  \textbf{Algorithm \thealgocounter:} #1\vspace{2pt}
  \hrule\vspace{4pt}
  \small\noindent
}{%
  \vspace{4pt}\hrule
  \bigskip
}
 
\usepackage{hyperref}
\hypersetup{
    colorlinks=true,
    linkcolor=red,
    filecolor=magenta,
    urlcolor=blue,
    citecolor=purple,
    pdftitle={CaLLM: Continual Adaptation of LLMs via Non-Parametric Adapter Routing},
    pdfpagemode=FullScreen,
}
 
\usepackage[capitalize,noabbrev]{cleveref}
\theoremstyle{plain}

\theoremstyle{definition}

\theoremstyle{remark}

\collasfinalcopy 
 
\title{CaLLM: Continual Adaptation of LLMs via Non-Parametric Adapter Routing}
 
\author{Bilge Celik \\
Eindhoven University of Technology \\
\texttt{b.celik@tue.nl}
\And
Anna Vettoruzzo \\
Eindhoven University of Technology \\
\texttt{a.vettoruzzo@tue.nl}
\And
Joaquin Vanschoren \\
Eindhoven University of Technology \\
\texttt{j.vanschoren@tue.nl}
}
 
\begin{document}
 
\maketitle
 
\begin{abstract}
Continual adaptation of large language models (LLMs) to evolving task distributions while mitigating catastrophic forgetting remains an open problem. Existing strategies often struggle to scale or lack a robust mechanism for zero-shot task-to-expert assignment.
We introduce CaLLM, a framework that resolves these tensions by combining metric-based meta-learning with modular parameter isolation via lightweight adapters. First, CaLLM projects each input batch into a task-agnostic embedding space to find its nearest prototype. It then routes the batch to the specific adapter specialized for that data distribution. 
This formulation provides (i)~controllable cross-task interference by design, (ii)~constant computational complexity regardless of task count, (iii)~seamless integration of novel distributions through prototype expansion, and (iv)~regime agnosticism as the local routing decision enables a unified approach to both online and offline settings. Our experiments demonstrate that this routing mechanism discovers semantically coherent task groupings while facilitating constructive interference between similar tasks, and keeping overhead at a minimum. Across diverse and fine-grained continual learning benchmarks, CaLLM consistently outperforms traditional isolation and replay methods, establishing prototype-based modular routing as a principled and scalable approach to lifelong adaptation of LLMs.
 
\end{abstract}
 
\section{Introduction}
\label{intro}
 
Large language models (LLMs) have become remarkable generalists, but real-world deployment often requires them to acquire task-specific expertise without sacrificing their broader capabilities. The same underlying model needs to be continually adapted to different use cases ranging from closely related to radically different, without retraining from scratch. However, the standard practice of sequential fine-tuning typically triggers catastrophic forgetting \citep{Kirkpatrick2017EWC}, where adaptation to a new domain erodes performance on previously learned ones. A recent survey \citep{Shi2025Survey} formalizes this challenge along two axes: vertical continuity, moving from broad pretraining to task-specific fine-tuning, and horizontal continuity, adapting to shifting tasks and data distributions over time. Without careful management, adaptation along either of these dimensions typically results in severe performance degradation.
 
This problem is particularly pronounced for aligned, instruction-tuned LLMs. \citet{TRACE2023} demonstrate that even parameter-efficient methods such as LoRA \citep{LoRA2021} can severely degrade an aligned model's general abilities and instruction-following quality after only a few rounds of sequential adaptation. Classical continual learning (CL) strategies such as regularization, experience replay, and orthogonal subspace isolation address specific aspects of this problem but they entail fundamental trade-offs. Regularization methods collapse when the model must transition between radically different domains (e.g., from natural language to code). Replay requires growing memory buffers that become unscalable as the task stream lengthens. 
Meanwhile, full retraining after each new task is computationally prohibitive, and real-world data often arrives as a continuous, non-stationary stream rather than the neatly delineated task blocks assumed in most CL benchmarks. Ideally, CL methods should adapt efficiently to new tasks in both online and offline settings.
 
We present CaLLM (Continual Adaptation of LLMs), a modular CL method that sidesteps these limitations through two interlocking components. First, CaLLM maintains a pool of modular experts (e.g., lightweight adapters implemented as LoRAs), each learning to specialize to a distinct region of the observed input distribution, ensuring that new knowledge is internalized without interfering with prior knowledge. Second, a lightweight non-parametric router, inspired by Prototypical Networks \citep{Snell2017ProtoNet}, maps each incoming batch to its most relevant adapter by comparing its (frozen-backbone) embeddings against stored prototypes. This mechanism dynamically and efficiently manages the adapter pool: it spawns new adapters when the data is sufficiently novel and re-purposes the least-used modules when the pool reaches full capacity. While Prototypical Networks use class-mean embeddings for few-shot classification, CaLLM uses distributional prototypes for task-level routing. This approach inherits the non-parametric simplicity of metric learning, where no gradient-based router optimization is required, and the model absorbs novel distributions simply by initializing new prototypes. A core architectural advantage of CaLLM is its constant computational overhead: only one adapter is active per forward pass, while the remaining adapters reside on disk.
 
We evaluate CaLLM on two complementary benchmarks: TRACE-8, which tests robustness to radical cross-domain shifts, and CITB-19\footnote{We refer to this benchmark as CITB \citep{CITB2023} after the originating paper; specifically utilizing the \textit{InstrDialog} task stream with additional synthetic data as specified in Appendix~\ref{app:synthetic_data}.}, which tests scalability on a long horizon dialogue stream, under offline (task-incremental) and online protocols. CaLLM outperforms CL baselines on the majority of tasks in both benchmarks, with the largest gains observed on specialized domains where the pretrained backbone is weakest. Furthermore, a systematic analysis of pool capacity reveals that moderate pooling outperforms both full isolation and single-adapter training in online settings. In this regime, we observe that interleaved gradients from semantically grouped tasks function as implicit replay, fostering constructive interference that enhances generalization without the overhead of a physical replay buffer.

In summary, our primary contribution is the complete CaLLM system, which unifies non-parametric routing with dynamic adapter isolation. Specifically:
\begin{itemize}
    \item We design and validate an end-to-end modular framework that achieves robust performance while maintaining a constant computational overhead by activating only one adapter per forward pass.
    \item We introduce a training-free routing mechanism that scales dynamically to multiple tasks without requiring gradient-based router optimization.
    \item We provide a comprehensive evaluation under both task-incremental and long-horizon online streaming scenarios, uncovering how constrained pool capacities foster constructive cross-task transfer. 
\end{itemize}
 
The remainder of this paper is organized as follows. Section §\ref{related} contextualizes our work in the CL literature, before section §\ref{callm} details the CaLLM method. Section §\ref{sec:results} presents our empirical analysis, including router representations, task-incremental and online results, and pool capacity effects. Section §\ref{conclusions} summarizes our findings and future directions.
 
\section{Related Work}
\label{related}
 
\paragraph{Continual learning foundations.} CL aims to accumulate knowledge from non-stationary streams without catastrophic forgetting. In this paradigm, each newly introduced task represents an inherently out-of-distribution shift relative to the historical training data seen by the model so far. Existing methods generally utilize replay \citep{Rolnick2019ER,LopezPaz2017GEM}, regularization \citep{Farajtabar2020OGD}, or parameter isolation \citep{Rusu2016ProgressiveNets, kang2023forget}. Beyond architectural design, evaluation is a critical aspect. Recent studies advocate for per-iteration metrics to detect the ``stability gap"---a transient but severe forgetting phase during training \citep{DeLange2023StabilityGap}---and transferability-aware embeddings to leverage inter-task relationships \citep{Wu2025TaskEmbed}. Online Continual Learning (OCL) \citep{Cai2021OnlineCL} further emphasizes learning efficacy via a single-pass ``test-then-train" protocol. This regime requires optimizing for immediate adaptation rather than long-term convergence, tracking performance through temporal moving averages instead of static end-of-task scalars.
 
\paragraph{Continual learning for LLMs.} Adapting LLMs involves navigating vertical (pretrain-to-fine-tune) and horizontal (temporal shift) continuity. We focus on Continual Fine-Tuning (CFT), evaluated via benchmarks like TRACE (cross-domain robustness) and CITB (long-horizon instruction streams). LoRA is the primary vehicle for CFT, with methods managing interference via subspace isolation, enforcing orthogonality through geometric constraints \citep{Wang2023OLoRA, Liang2024InfLoRA} or gradient projection \citep{Qiao2024LBCL}. TreeLoRA~\citep{qian2025treelora} organises per-layer LoRAs in a gradient-similarity tree, exploring task relationships via lower-confidence-bound bandits. However, these often sum branches with equal weight at inference, failing to select the most relevant parameters. While GainLoRA \citep{Liang2025GainLoRA} and SwitchCIT \citep{Wu2024SwitchCIT} introduce gating or switch networks, they require specialized optimization. Conversely, replay-based methods \citep{Scialom2022CT0, Wang2024InsCL, He2024SEEKR} maintain performance but add memory overhead. Notably, recent work suggests that much degradation in LLMs stems from disrupted task alignment rather than true knowledge loss \citep{Chen2025Spurious}.
 
\paragraph{Routing and mixture-of-experts approaches.} Sparse Mixture-of-Experts (MoE) architectures decouple model capacity from per-input computation by routing each token to a small subset of parallel experts \citep{Fedus2022Switch, Jiang2024Mixtral}. Learned routers discover interpretable data partitions aligned with latent clusters \citep{Dikkala2023RoutingBenefits}. In parameter-efficient fine-tuning (PEFT), LoRAMoE \citep{Dou2024LoRAMoE} and MoRAL \citep{Yang2024MoRAL} use balancing losses to preserve world knowledge, while MoA \citep{Feng2024MixtureLoRAs}, SEE \citep{Wang2025SEE}, and MLLM-CL \citep{MLLMCL2025} assigns specific LoRA modules to different tasks or modalities, routing them at inference time to minimize interference. Despite their modularity, these methods generally rely on gradient-based gating and specialized training objectives and they are highly susceptible to routing drift and catastrophic forgetting within the router itself. Furthermore, they require complex balancing and regularization losses to prevent expert collapse. CaLLM overcomes this by using a non-parametric metric-based router, effectively decoupling router scaling from gradient optimization.
 
\paragraph{Prototypical networks and metric-based routing.}CaLLM’s router is grounded in metric-based meta-learning \citep{Vettoruzzo2024MetaLearning}, drawing inspiration from Prototypical Networks. By computing distances to centroids in a learned embedding space, this non-parametric bias enables robust adaptation without inner-loop optimization. While prototypes have traditionally been used for low-level class-incremental classification \citep{DeLange2021ProtosCL, Zhu2021ProtoCL}, CaLLM transposes this to high-level system routing. Rather than using embeddings to represent a single class, CaLLM uses embedding prototypes to capture the geometric boundaries of entire multi-task streams. This allows the model to map task-level representation to isolated parameter spaces that are assigned to specific adapters. This yields a training-free router that inherits the modularity of MoE but avoids the overhead of learned gating. By combining parameter isolation with an online, no-replay constraint, CaLLM provides a scalable alternative that handles novel tasks via prototype expansion and bounds memory through an repurposing policy.

\section{Continual Adaptation of LLMs (CaLLM)}
\label{callm}

\begin{figure}[t]
\centering
\includegraphics[width=0.9\linewidth]{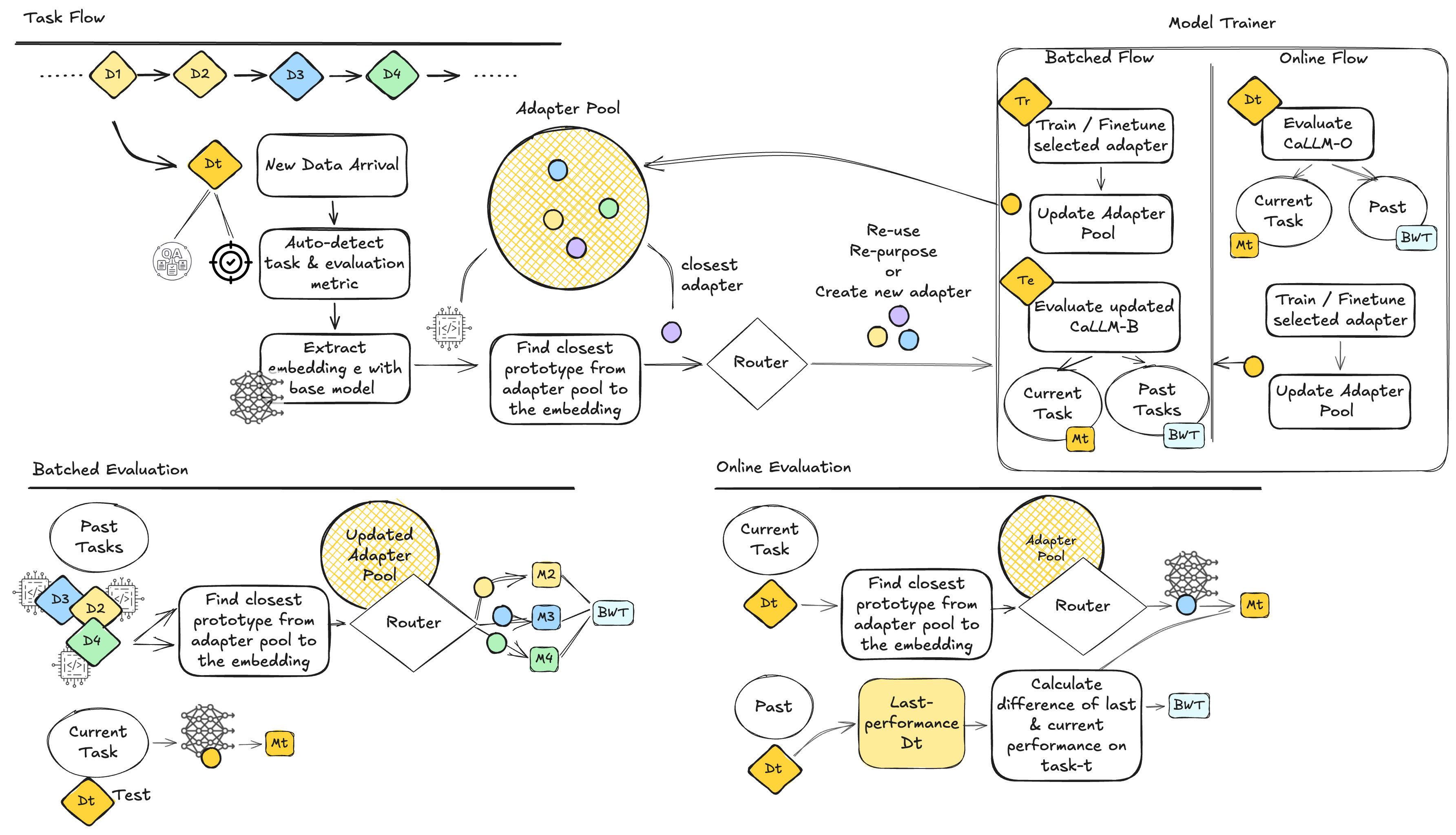}
\caption{\textbf{CaLLM system overview (training + evaluation).} Each new data batch is embedded with a frozen base model (using mean-pooling). The non-parametric router manages a dynamic pool of specialized adapters, each mapped to a prototype. Based on semantic similarity between the new embedding and existing prototypes, the router triggers the reuse, creation, or repurposing of an adapter. The selected adapter is then trained or finetuned and the pool is updated. The diagram further details how this routing mechanism integrates with the distinct training and evaluation loops.}
\label{fig:method_overview}
\end{figure}

CaLLM addresses catastrophic forgetting through a prototype-based routing mechanism coupled with dynamic, task-specific parameter-efficient adapters. The central idea is to maintain a pool of lightweight adapter modules, each specializing in a region of the task embedding space. As illustrated in Figure~\ref{fig:method_overview}, the system pipeline integrates four primary processes: (i) a non-parametric router that extracts embeddings from incoming data and maps data batches to adapters; (ii) an adapter pool that manages a collection of $P$ parameter-efficient adapters and their associated prototypes; (iii) a model trainer that performs localized SFT on selected adapters; and (iv) an evaluation pipeline that assesses performance across task streams to quantify learning progress and monitor for catastrophic forgetting.
By routing data based on semantic similarity or spawning new adapters for sufficiently novel inputs, CaLLM decouples task capacity, allowing selective reuse or isolation as the task distribution evolves.
 
CaLLM operates in two modes: \emph{Batched} (CaLLM-B), where tasks arrive sequentially, and \emph{Online} (CaLLM-O), where small, randomly interleaved chunks from multiple tasks arrive in a non-stationary stream. Both share routing, training, and evaluation components but differ in data presentation order, evaluation protocol, and CL metric computation.

\subsection{Adapter Pool}
\label{sec:adapter_pool}

CaLLM maintains a pool $\mathcal{P} = \{(\ell_i, \mathbf{p}_i, c_i)\}_{i=1}^{|\mathcal{P}|}$ of parameter-efficient adapters, with maximum capacity~$P$. Each adapter $\ell_i$ is paired with a prototype embedding $\mathbf{p}_i \in \mathbb{R}^d$ that summarizes the data distribution it was trained on and a usage counter~$c_i$. The proposed system is both backbone- and adapter-agnostic, supporting any Transformer-based LLM and any PEFT method that can be independently instantiated, saved, and loaded (e.g., LoRA \citep{LoRA2021}, $(IA)^3$ \citep{IA32022}, or adapter modules \citep{Houlsby2019Adapters}). We use the terms \emph{expert} and \emph{adapter} interchangeably here, since each expert in the pool is realized as a parameter-efficient adapter. In our experiments we use LoRA as the default adapter, where each $\ell_i$ consists of low-rank decompositions $\Delta W = BA$ injected into all linear projections in each Transformer block, parameterized with rank~$r$ and scaling factor $\alpha = 2r$. The modified forward pass for a projection $W$ becomes:
\begin{equation}
    \mathbf{h}' = W\mathbf{x} + \frac{\alpha}{r} B_i A_i \mathbf{x},
    \label{eq:adapter_forward}
\end{equation}
where $A_i \in \mathbb{R}^{d_{\text{in}} \times r}$ and $B_i \in \mathbb{R}^{r \times d_{\text{out}}}$ are the trainable low-rank factors of the currently loaded adapter~$\ell_i$. All backbone parameters remain frozen throughout training and only the active adapter’s parameters are updated. This ensures that (i)~the base model’s representations remain stable, and (ii)~adapters from different tasks do not interfere with one another at the parameter level. Because only one adapter is active per forward pass while other reside in secondary storage, CaLLM ensures \emph{fixed-cost modularity}: training FLOPs and GPU memory remain constant regardless of pool size $|\mathcal{P}|$. Notably, routing does not compromise this property. Its additional pass through the frozen backbone is encoder-only, reuses the already-loaded model, and is invoked once per task (or once per chunk in the online setting) rather than once per query. Consequently, routing contributes only about 0.5\% of training FLOPs and less than 0.1\% of per-query inference cost (see Appendix~\ref{app:compute} for quantitative results).

\subsection{Non-Parametric Router}
\label{sec:router}

The router maps incoming data to the most appropriate adapter in the pool via two stages: embedding extraction and prototype matching.

\paragraph{Embedding extraction.}Given a batch of $n$ input prompts, the router extracts a distributional representation $e$ by performing a forward pass through the frozen backbone with all adapters disabled. Let $\mathbf{h}_{jt}^{(L)}$ denote the hidden state at position~$t$ of sequence~$j$ in the final transformer layer~$L$. We compute a single embedding by mean-pooling over valid (non-padding) tokens and averaging across the batch:
\begin{equation}
    \mathbf{e} = \frac{1}{n} \sum_{j=1}^{n} \frac{\sum_{t=1}^{s_j} m_{jt} \cdot \mathbf{h}_{jt}^{(L)}}{\sum_{t=1}^{s_j} m_{jt}},
    \label{eq:embedding}
\end{equation}
where $m_{jt} \in \{0, 1\}$ is the attention mask and $s_j$ is the sequence length. Input sequences are tokenized with truncation at a configurable maximum length. By using the frozen backbone without adapters, the embeddings reflect the base model’s representation of the input distribution, providing a stable reference frame that works from $t=0$ and does not drift as adapters are trained.

\paragraph{Prototype matching.}The router computes the cosine distance between the new embedding $e$ and each stored prototype $p_i$:
\begin{equation}
    d_i = 1 - \frac{\mathbf{e} \cdot \mathbf{p}_i}{\|\mathbf{e}\| \, \|\mathbf{p}_i\|}, \quad i = 1, \ldots, |\mathcal{P}|.
    \label{eq:cosine_dist}
\end{equation}
The adapter selection then follows a three-way decision rule governed by a distance threshold~$\eta$:

\begin{enumerate}
    \item \textbf{Reuse} ($d_{\min} \leq \eta$): Load the closest adapter $\ell_{\arg\min_i d_i}$, update its prototype via an exponential moving average with equal weighting $\mathbf{p}_i \leftarrow \tfrac{1}{2}(\mathbf{p}_i + \mathbf{e})$, and increment its usage counter~$c_i$.
    \item \textbf{Create} ($d_{\min} > \eta$ and $|\mathcal{P}| < P$): Initialize a new adapter from the base weights, add it to the pool with prototype $\mathbf{p}_{\text{new}} = \mathbf{e}$ and usage counter $c_{\text{new}} = 1$.
    \item \textbf{Repurpose} ($d_{\min} > \eta$ and $|\mathcal{P}| = P$): Repurpose the least-used adapter slot (by usage count $c_i$), reassign its prototype to $\mathbf{e}$, reset its usage counter, and overwrite its weights during subsequent fine-tuning. Although repurposing can overwrite a rare adapter, our capacity analysis shows the router consistently groups similar tasks together, preventing such loss in practice.
\end{enumerate}

During evaluation, the router operates in \emph{read-only mode}. It finds the closest prototype and returns the associated adapter without creating or modifying prototypes. The threshold $\eta$ controls the granularity of the pool, lower values encourage more adapters (finer specialization), while higher values promote sharing across related distributions. We analyze the effect of pool capacity in \S\ref{sec:pool_capacity}.

When no fixed threshold is provided, an adaptive threshold is computed as half the minimum pairwise cosine distance between the closest prototype and all others:
\begin{equation}
    \eta_{\text{adapt}} = \frac{1}{2} \min_{j \neq i^*} \left(1 - \frac{\mathbf{p}_{i^*} \cdot \mathbf{p}_j}{\|\mathbf{p}_{i^*}\| \, \|\mathbf{p}_j\|}\right),
    \label{eq:adaptive_threshold}
\end{equation}
where $i^* = \arg\min_i d_i$ is the closest prototype to the incoming embedding. This ensures the threshold tightens as the embedding space becomes more densely populated. When only one prototype exists, a default threshold of $0.1$ is used.

\subsection{Training and Evaluation Protocol}
\label{sec:training_procedure}

Each training step applies SFT on the selected adapter using the standard causal language modeling cross-entropy loss:
\begin{equation}
    \mathcal{L} = -\sum_{t=1}^{|\mathbf{y}|} \log P_{\theta_{\text{adapt}}}(y_t \mid y_{<t}, \mathbf{x}),
    \label{eq:sft_loss}
\end{equation}
where $\mathbf{x}$ is the input prompt, $\mathbf{y} = (y_1, \ldots, y_{|\mathbf{y}|})$ is the target sequence, and $\theta_{\text{adapt}}$ denotes the adapter parameters. Optimization hyperparameters (e.g. optimizer, learning rate schedule, gradient clipping, numerical precision) as well as input formatting are all configurable and decoupled from the core method. Specific settings used in our experiments are reported in \S\ref{sec:results}.

CaLLM supports two CL regimes that share the training objective above but differ in data presentation, evaluation timing, and metric computation. Full pseudocode for both protocols is provided in Appendix~\ref{app:algorithms}.

\textbf{Batched task-incremental protocol.}
In Batched mode, tasks arrive sequentially as complete datasets $\mathcal{D}_1, \mathcal{D}_2, \ldots, \mathcal{D}_T$, each with train, evaluation, and test splits (Algorithm~\ref{alg:batched}). For each task with dataset~$\mathcal{D}_t$, the system first extracts an embedding from a sample of the training split and queries the router (with prototype updates enabled) to select or create an adapter. The selected adapter is then loaded onto the base model and fine-tuned on the full training split using the SFT objective (Eq.~\ref{eq:sft_loss}), with the evaluation split used for early stopping and checkpoint selection. After training, the adapter is saved to disk.

Evaluation proceeds in two phases. First, the trained adapter is evaluated on the current task's test set to measure adaptation quality. Second, to assess backward transfer, all previously seen tasks $j < t$ are re-evaluated: for each prior task, the router operates in read-only mode (no prototype updates) to find the best-matching adapter under the current prototype set, and that adapter is loaded and evaluated on task~$j$'s test set. Backward transfer is then computed as the average normalized performance change across prior tasks.

\textbf{Online continual learning protocol.}
In Online mode, data arrives as a stream of small chunks randomly interleaved across multiple tasks in a non-stationary distribution. (Algorithm~\ref{alg:online}). Each chunk contains a fixed number of examples ($chunk\_size$) from a single task.

Following the \emph{test-then-train} paradigm, the protocol begins with the router extracting a chunk's embedding $e_i$ to select, create, or repurpose an adapter according to the logic described above. Once the parameter subspace is determined, the chunk is immediately evaluated using the selected adapter, or the frozen backbone if a new adapter is being initialized, prior to any gradient updates. This pre-update score provides an unbiased measure of the system's current knowledge and is used to update the continual learning metrics defined below.
After evaluation, the selected adapter is fine-tuned on the input chunk via SFT (Eq.~\ref{eq:sft_loss}). This single-pass procedure ensures structural consistency and strictly avoids post-update evaluation to prevent information leakage, allowing CaLLM to satisfy the core assumptions of online learning while tracking knowledge accumulation in real-time.

\textbf{Continual learning metrics.}
We report the following measures to quantify plasticity and stability across the task stream:

\begin{itemize}
    \item \textbf{Average Performance:} In the batched setting, after learning task $t$, we compute the normalized average over all tasks seen so far (each evaluated with its best-matching adapter via read-only routing): $\text{Avg}_t = \tfrac{1}{t}\sum_{j=1}^{t} R_{j,t}/s_j$, where $R_{j,t}$ is the performance on task $j$ after learning task $t$ and $s_j$ is the metric scale. This captures overall knowledge retention and adaptation quality at each point in the task sequence. 
    
    In the online setting, the equivalent is the \textbf{Online Moving Average (OMA)}, a running normalized average updated at each step with the latest pre-update evaluation: $\text{OMA}_t = \tfrac{1}{t}\sum_{k=1}^{t} R_k^{\text{pre}}/s_k$, where $s_k$ is the task-specific metric scale at step~$k$. As a moving average over the stream, OMA tracks how the system's overall performance evolves in real-time, reflecting its ability to generalize to new distributions.
    \item \textbf{Backward Transfer (BWT):} Measures the impact of learning new tasks on previously learned ones. In the batched setting, after learning task $t$: $\text{BWT}_t = \tfrac{1}{t-1}\sum_{j=1}^{t-1}(R_{j,t} - R_{j,j})/s_j$, where $R_{j,j}$ is the performance immediately after learning task $j$ and $s_j$ is the metric scale. 
    
    In the online setting, BWT is computed as: $\text{BWT}_{j,t} = (R_j^{\text{pre},t} - R_j^{\text{pre},\text{prev}}) / s_j$, where $R_j^{\text{pre},\text{prev}}$ is the pre-update score from the most recent prior observation of task~$j$. Unlike the batched variant above, online BWT measures the change between consecutive observations of the same task rather than from the initial post-training score.
\end{itemize}

\section{Experiments \& Results}
\label{sec:results}

We organize our experimental analysis around a progression from defining setup to understanding mechanism, followed by performance evaluation and stress-testing. 

\subsection{Experimental Setup}
\label{sec:setup}

\textbf{Benchmarks.} We evaluate CaLLM on two complementary benchmarks that test different axes of CL difficulty:

\begin{enumerate}
    \item \textbf{TRACE-8 \citep{TRACE2023} (Diversity \& Shift)} comprises 8 heterogeneous tasks spanning domain-specific QA (ScienceQA, FOMC, MeetingBank), multilingual processing (C-STANCE, 20Minuten), code completion (Py150), and math reasoning (NumGLUE-cm/ds) testing robustness to radical distribution shifts. We utilize the variant with $5000$ training examples per task.

    \item \textbf{CITB-19 \citep{CITB2023} (Scale \& Granularity)} consists of 19 dialogue-centric tasks derived from Super-NaturalInstructions \citep{wang2022supernaturalinstructions}, categorized into Dialogue State Tracking (4), Dialogue Generation (11), and Intent Identification (4). This suite is used to test long-horizon streams and fine-grained intra-domain discrimination. To ensure training stability, we utilize a refined version of this benchmark augmented with $3000$ high-quality synthetic examples per task (see Appendix~\ref{app:synthetic_data} for the generation and filtering protocol). This is the dataset version used in all our experiments.
\end{enumerate}

Task performance is measured using standard metrics appropriate to each task type: accuracy for classification and multiple-choice tasks, ROUGE-L for summarization and all CITB-19 tasks, SARI for text simplification, and EDIM (edit-distance fuzzy match ratio) for code completion. In label-free stream settings, metric assignment is handled automatically via LLM-based task classification (Appendix~\ref{app:auto_metric}).

All input data is formatted using an Alpaca-style instruction template (\texttt{Instruction / Response / <eos>}), ensuring a uniform interface across diverse task types. Following the protocols defined in \S\ref{sec:training_procedure}, the \textit{Batched} setting processes tasks as complete sequential blocks, while the \textit{Online} setting uses a non-stationary stream of $max\_chunks$ steps. Each step consists of $chunk\_size$ examples, with chunks randomly interleaved across tasks using a fixed seed to ensure reproducibility.

\textbf{Baselines.} We compare CaLLM against five representative CL strategies:
~\textbf{Sequential Adapter Fine-tuning (Seq-FT)}, a single LoRA adapter fine-tuned sequentially on the task stream;
~\textbf{Inference-only}, the zero-shot performance of the frozen backbone;
~\textbf{Elastic Weight Consolidation (EWC)} \citep{Kirkpatrick2017EWC}, a regularization-based method penalizing shifts in important weights;
~\textbf{Experience Replay (ER)} \citep{Rolnick2019ER}, a replay-based method with a fixed memory buffer;
~\textbf{Orthogonal LoRA (oLoRA)} \citep{Wang2023OLoRA} a parameter-efficient CL baseline that learns tasks in orthogonal low-rank subspaces.
~\textbf{TreeLoRA}~\citep{qian2025treelora}: layer-wise LoRAs organised by a hierarchical gradient-similarity tree with bandit-based task-relationship exploration.

All experiments use gemma-3-12b-it (4-bit quantized) as the backbone, selected for its superior zero-shot performance across all tasks (see Appendix~\ref{app:backbone}). Appendix~\ref{app:implementation} shows implementation details and hyperparameter configuration. The code will be openly released upon paper acceptance.

\subsection{Router Representations: Why Modular Routing Works}
\label{sec:router_mechanism}
\begin{wrapfigure}{r}{0.4\textwidth}
\vspace{-5mm}
  \begin{center}
    \includegraphics[width=0.4\textwidth]{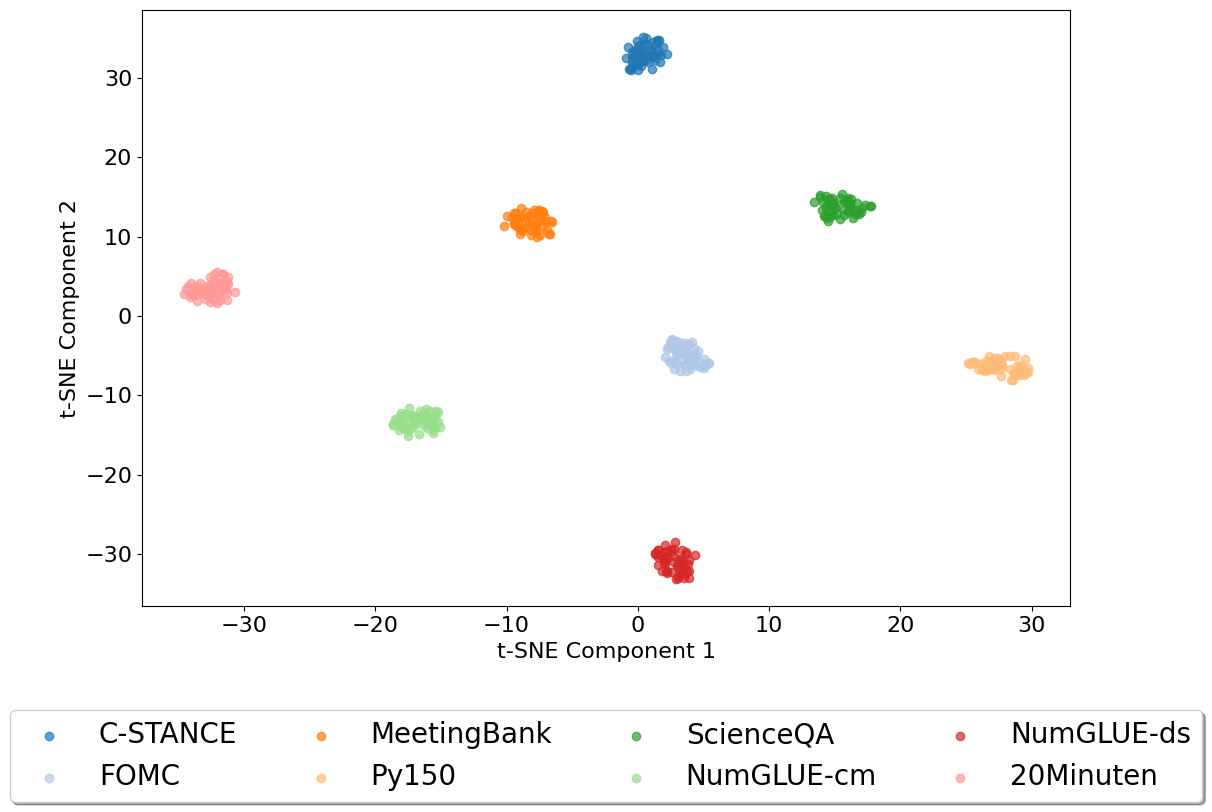}
  \end{center}
  \caption{\textbf{Router Embedding Space (2D Projection) for TRACE-8.} t-SNE projections of the router's embeddings, colored by task identity.} \label{fig:router_embeddings}
\vspace{-4mm}
\end{wrapfigure}
CaLLM's ability to mitigate interference in task-agnostic settings depends on the router's capacity to recognize distributional boundaries from input features alone.
Figure~\ref{fig:router_embeddings} visualizes 2D projections of the TRACE-8 benchmark. The router produces well-separated clusters, with radical domain shifts (e.g., code vs.\ multilingual text) pushed far apart in embedding space. This geometric separation is expected given the large distributional gaps between TRACE tasks, but it confirms that the router’s lightweight architecture and a base-model representation is sufficient to capture these boundaries without additional finetuning.

Similarly, on CITB-19 (Figure~\ref{fig:router_embeddings_complete}(b) in Appendix~\ref{app:proto_embeddings}), the router discriminates between 19 fine-grained dialogue sub-tasks, such as dialogue state tracking and intent identification, despite their shared conversational format. This intrinsic geometric separation allows for reliable, task-agnostic routing. When pool capacity $P$ is sufficient, each cluster maps to a dedicated adapter, achieving full parameter isolation. The consequences of restricting $P$ and forcing cluster merges are analyzed in \S\ref{sec:pool_capacity}.

\subsection{Task-Incremental Results}
\label{sec:batched_results}

We first evaluate CaLLM’s end-to-end performance in the standard batched task-incremental setting, where tasks arrive one at a time and the system trains on each complete dataset sequentially. 

Final performance by individual task for CALLM-B and other baselines for TRACE benchmark is shown in Table~\ref{tab:trace_per_task}. CaLLM-B achieves the highest score among all baseline methods on 4 of 8 tasks. The most significant gains over the inference-only baseline appear on specialized domains where the pre-trained backbone is weakest: code completion (Py150, $+2.33$ in EDIM), science reasoning (ScienceQA, $0.35 \to 0.81$ in accuracy), and math (NumGLUE-ds, $6\times$ improvement in accuracy). TreeLoRA's wins concentrate on the two arithmetic splits, where its hierarchical gradient-similarity tree consolidates closely related skills (mean accuracy $0.60$ across NumGLUE-cm/ds, versus $0.28$ for CaLLM-B), but it collapses on Py150 ($5.52$ vs.\ $17.24$ in EDIM), where the code-generation gradient signature is dissimilar from the natural language tasks and the tree potentially mis-routes the adapter, a failure mode that CaLLM-B's prototype router avoids by treating each task as a first-class cluster. Conversely, on C-STANCE and 20Minuten, performance is largely inherited from the backbone; on the former, fine-tuning slightly degrades pre-existing competence, while on the latter, all methods fall within a narrow 2-point SARI band. Because 20Minuten is the final task in the sequence, no method suffers from forgetting, narrowing the performance gap between CaLLM-B and baselines.
Notably, CL baselines underperform Inference-Only on several tasks (e.g., C-STANCE, FOMC, MeetingBank, Py150). We attribute this to Inference-Only's strong large-scale backbone---whose pre-trained competence makes it a strong baseline rather than a trivial lower bound---and to its avoidance of the catastrophic forgetting that erodes CL baselines as the task stream grows (Appendix \ref{app:baselines} shows this gap persists even with per-baseline hyperparameter tuning). CaLLM-B avoids this trade-off by routing task-specific updates to isolated experts, preserving the backbone's zero-shot competence while enabling specialization.

\begin{table}
  \centering
  \caption{\textbf{Per-Task Final Performance on TRACE-8 (Batched).}
  Final scores after training on all tasks sequentially.}
  \label{tab:trace_per_task}
  \small
  \begin{tabular}{llccccccc}
    \toprule
    \textbf{Task} & \textbf{Metric}
      & \textbf{Inf-Only} & \textbf{Seq-FT} & \textbf{EWC}
      & \textbf{ER} & \textbf{oLoRA} & \textbf{TreeLoRA \footnote{\footnote{TreeLoRA is reported on TRACE-8 only, as its public implementation is TRACE-oriented and its cumulative checkpoint-wise inference protocol would make evaluation over the 19-task CITB stream computationally prohibitive.}}} & \textbf{CaLLM-B} \\
    \midrule
    C-STANCE    & acc (0--1)      & \textbf{0.62}  & 0.03  & 0.28  & 0.04  & 0.09  & 0.41  & 0.43 \\
    FOMC        & acc (0--1)      & 0.64  & 0.30  & 0.31  & 0.14  & 0.06  & 0.51  & \textbf{0.68} \\
    MeetingBank & ROUGE-L (0--1)  & 0.21  & 0.06  & 0.10  & 0.10  & 0.16  & 0.11  & \textbf{0.23} \\
    Py150       & EDIM (0--100)   & 14.91 & 11.96 & 10.55 & 12.98 & 9.81  & 5.52  & \textbf{17.24} \\
    ScienceQA   & acc (0--1)      & 0.35  & 0.13  & 0.61  & 0.19  & 0.00  & 0.72 & \textbf{0.81} \\
    NumGLUE-cm  & acc (0--1)      & 0.05  & 0.14  & 0.12  & 0.06  & 0.04  & \textbf{0.67} & 0.15 \\
    NumGLUE-ds  & acc (0--1)      & 0.07  & 0.28  & 0.15  & 0.13  & 0.05  & \textbf{0.54} & 0.42 \\
    20Minuten   & SARI (0--100)   & 39.22 & 38.06 & 40.06 & 39.63 & 39.91 & \textbf{41.05} & 39.49 \\
    \midrule
    Avg. perf. & normalized (0--1)
        & 0.29 & 0.16 & 0.26
        & 0.15 & 0.15
        & 0.39 & \textbf{0.41} \\
    BWT & (higher is better)
        & -- & -0.21 & -0.10
        & -0.18 & -0.11
        & -- & \textbf{0.01} \\
    \bottomrule
  \end{tabular}
\end{table}
\begin{figure}[h]
\begin{center}
\includegraphics[width=0.45\linewidth]{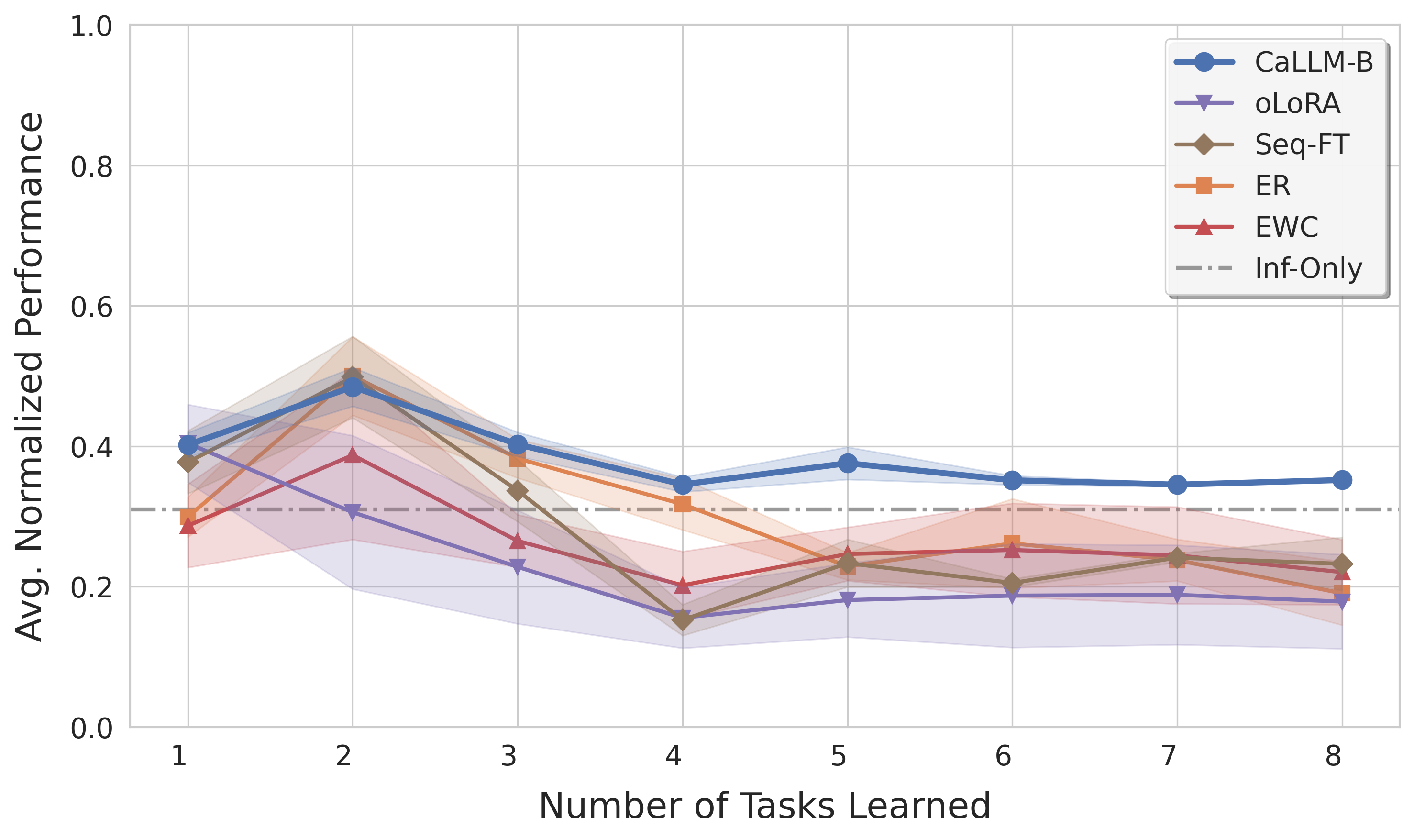}
\includegraphics[width=0.45\linewidth]{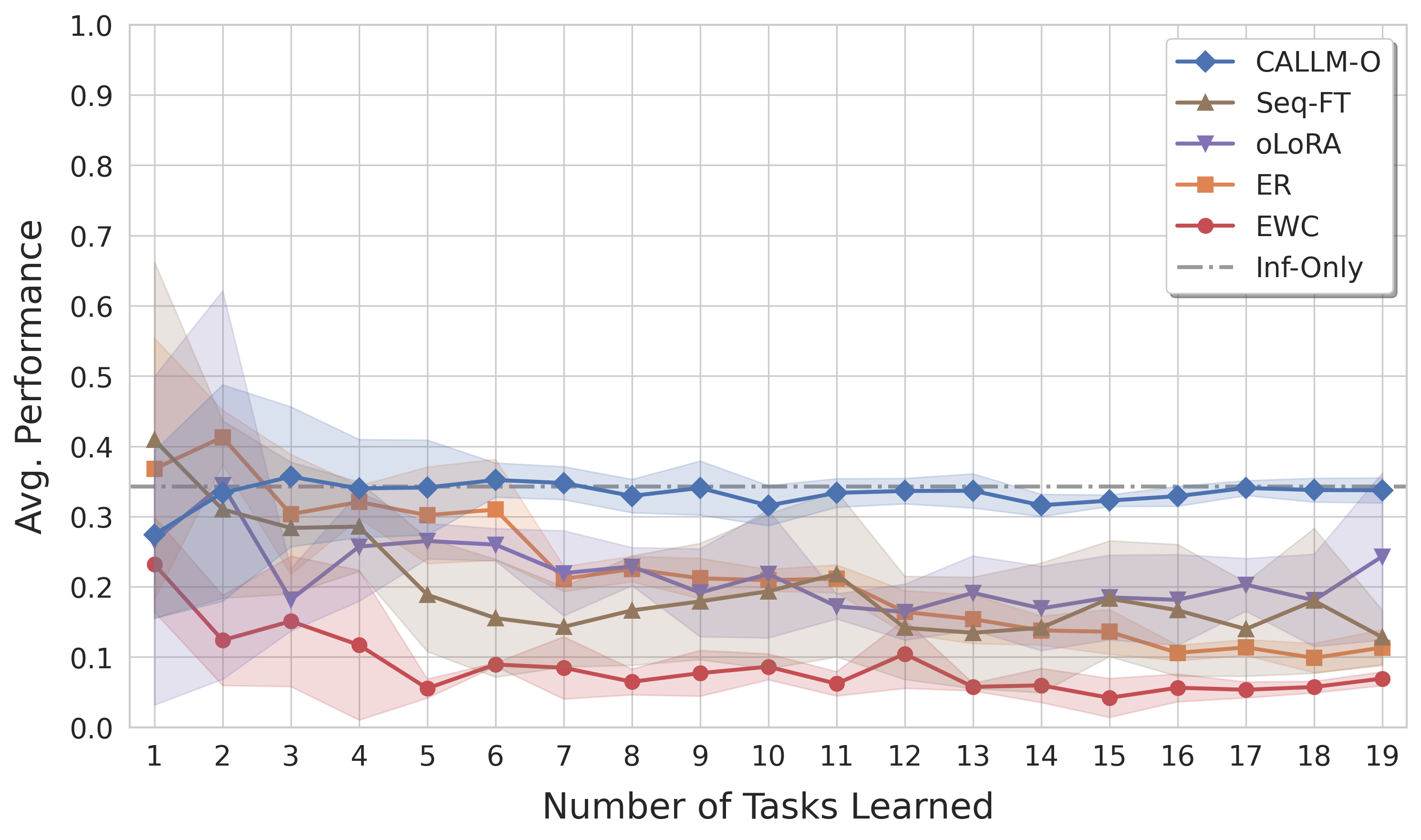}
\end{center}
\caption{\textbf{Task-Incremental Learning Performance (mean $\pm$1 std).} Average normalized score as tasks are added sequentially, for CaLLM-B and all baselines. \textbf{(a)}~TRACE-8. \textbf{(b)}~CITB-19.}
\label{fig:batched_baseline_comparison}
\end{figure}

Figure~\ref{fig:batched_baseline_comparison} shows the evolution of average normalized per-task performance across the sequential task stream both for TRACE-8 and CITB-19 datasets, while a detailed per-task performance breakdown can be seen in Tables~\ref{tab:pertask_trace_batched} and~\ref{tab:pertask_citb_batched} in Appendix~\ref{app:pertask_batched}. On TRACE-8, baselines struggle with radical distribution shifts. EWC's performance declines significantly upon encountering tasks that require more complex reasoning (\textit{Py150} at Task-4), while ER and oLoRA exhibits a steady downward trend in later tasks. In contrast, CaLLM-B demonstrates robust stability throughout the entire sequence, by routing incompatible distributions (e.g., code vs.\ math) to separate experts, as shown in Figure~\ref{fig:router_embeddings}(a). CaLLM-B finishes the 8-task sequence with an average normalized score of 0.44, substantially outperforming the Inference-Only baseline (0.31) and ER (0.31). We additionally verified that using Gemma~3's native chat template for the
Inference-Only baseline does not yield a stronger competitor,
making our
reported Inf-Only a conservative comparison
(Appendix~\ref{app:prompt-sensitivity}).
 The narrow variance bands ($\pm 1$ SD) confirm CaLLM-B's robustness to training stochasticity.
On CITB-19 (Fig.~\ref{fig:batched_baseline_comparison}b), the challenge shifts to sequence length. While Seq-FT degrades rapidly due to interference within a single adapter, CaLLM-B maintains a stable trajectory across all 19 tasks. By instantiating new experts as needed ($P \approx 19$), the system prevents parameter overloading. The tight variance bands across all 19 steps confirm that this stability is not degrading even when sequence task length increases. CaLLM-B achieves the highest score among CL methods on 12 of 19 tasks, demonstrating broad competence across fine-grained CITB dialogue subtypes (see Table~\ref{tab:citb_per_task}, Appendix~\ref{app:citb_per_task}). However, fine-tuning on CITB-19 does not improve over Inference-Only for any method, likely because dialogue tasks are within the natural competence of a 12B instruction-tuned model (see Appendix~\ref{app:citb_overlap}).

We also analyzed the BWT dynamics across both benchmarks in Figure~\ref{fig:bwt_comparison} (Appendix~\ref{app:pertask_batched}) in the batched scenario. Across both TRACE-8 and CITB-19, CaLLM-B is the only method that consistently maintains a positive, or close to 0, BWT, with a very low variance across different random seeds. Conversely, all competing baselines exhibit a steady degradation in BWT as more tasks are introduced to the stream. This stark contrast demonstrates that CaLLM successfully learn new tasks without forgetting the previously acquired knowledge.

These gains come at negligible additional compute.Because only one adapter is active per forward pass, CaLLM consumes the same
$7339$~TFLOPs of training compute as single-adapter Seq-FT. The router adds only a small increment: Using the standard $2N$ FLOPs-per-token estimate with
$N \approx 12 \times 10^{9}$ for \texttt{gemma-3-12b-it}, a routing call over $n \approx 20$ prompts of $\sim\!10$ tokens costs $\approx 4.8$~TFLOPs, so across TRACE-8's eight tasks the router consumes $\approx 38.4$~TFLOPs (only $\approx 0.5\%$ of training). At inference this amortizes over all evaluation queries within a task, giving a per-query overhead $\ll 0.1\%$ (see Appendix~\ref{app:compute} for further details). 

\subsection{Online Streaming Dynamics}
\label{sec:online_results}

Having established CaLLM's task-incremental performance, we now examine the fine-grained dynamics of online learning: How does the router behave over time as the data stream unfolds, and how does expert specialization emerge without supervision?
Unlike the batched setting, we do not report multi-seed variance here expecting the OMA metric averaging over 500 evaluation steps to provide inherent statistical stability.

\begin{figure}[h]
\begin{center}
\includegraphics[width=0.85\linewidth]{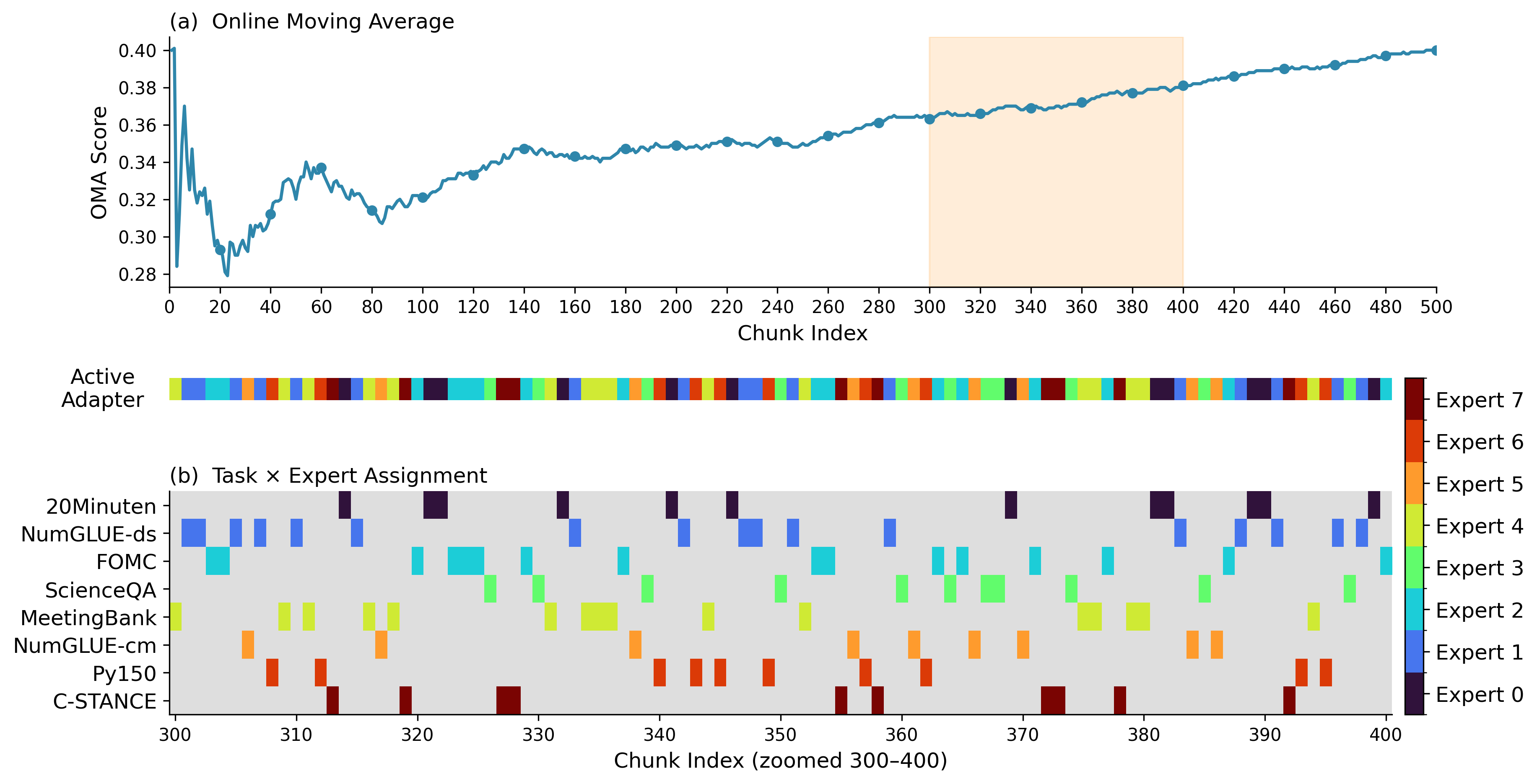}
\end{center}
\caption{\textbf{CaLLM Online Learning Dynamics (TRACE Stream, $P{=}8$).}
\textbf{(a)}~OMA.
\textbf{(b)}~Task $\times$ expert assignment heatmap; aggregate adapter timeline on top and per-task routing below.
See Appendix~\ref{app:online_heatmap} for the full 500-chunk view.}
\label{fig:online_trace}
\end{figure}

Figure~\ref{fig:online_trace} summarises CaLLM-O’s behavior over the TRACE online stream. The top panel (Figure~\ref{fig:online_trace}a) tracks the OMA performance whereas the bottom panel (Figure~\ref{fig:online_trace}b) zooms into a representative 100-chunk window (chunks~200--300) of the $task \times expert$ assignment heatmap, where each row is a task, each column a chunk, and color encodes the assigned expert. The full 500-chunk heatmap is provided in Appendix~\ref{app:online_heatmap} showing this behavior persists for the whole stream.

\textbf{Knowledge Accumulation.}
The OMA curve (Figure~\ref{fig:online_trace}a) rises steadily from ${\sim}0.29$ to ${\sim}0.40$ over the 500-chunk stream, confirming that CaLLM-O accumulates knowledge faster than it forgets, even in the absence of explicit task boundaries. After brief initial volatility during the first ${\sim}50$ chunks, where the router is still forming its expert assignments and initialized experts building up knowledge over tasks, the trajectory stabilizes into a monotonic upward trend, indicating that new tasks are absorbed without disrupting previously learned ones.

\textbf{Expert Specialization.}
The zoomed heatmap (Figure~\ref{fig:online_trace}b) reveals consistent task--expert matching within the displayed window. The top strip shows the aggregate adapter timeline with distinct color band showing how frequently the adapter changes through the stream. The per-task rows confirm that each task is reliably routed to the same expert when it reappears in the stream, indicating that the router functions as a distribution-sensitive switch rather than randomly cycling through experts.

\subsection{The Role of Pool Capacity}
\label{sec:pool_capacity}

The router embedding analysis (\S\ref{sec:router_mechanism}) showed that the router is capable of separating task clusters both in diverse and fine-grained task distributions. When pool capacity is unconstrained, each cluster maps to a unique expert, achieving full isolation. But in practice, memory budgets are limited and the number of tasks to be seen is not always known in advance. We now investigate what happens as pool size $P$ is progressively reduced, revealing a spectrum of routing behaviors from perfect isolation to system collapse.

\subsubsection{Expert Assignment Dynamics in Task-Incremental Learning}
\label{sec:capacity_dynamics}

\begin{figure}[h]
\begin{center}
\includegraphics[width=0.9\linewidth]{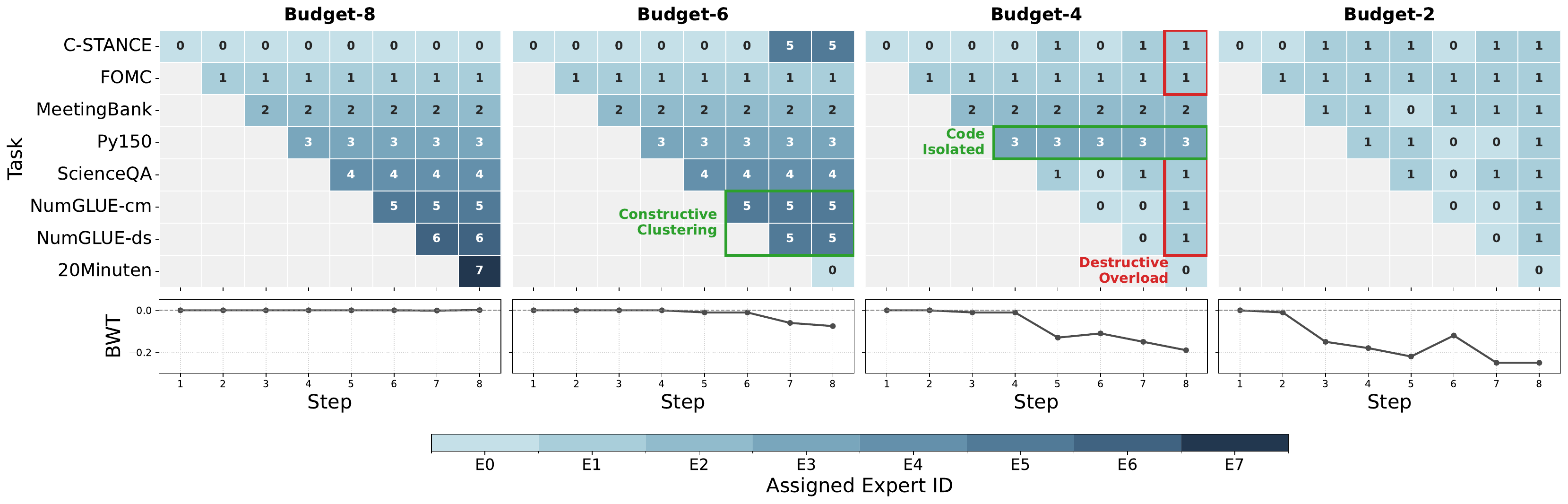}
\end{center}
\caption{Router behavior under pool capacity constraints for TRACE-8 in Batched scenario.}
\label{fig:router_time}
\end{figure}

Figure~\ref{fig:router_time} visualizes the evolution of expert assignments and the corresponding BWT curves on TRACE-8 as pool size varies from $P=8$ to $P=2$ (see Appendix~\ref{app:proto_embeddings} for extended plots).
\begin{itemize}
\item \textbf{Ideal Isolation ($P=8$):} The router assigns unique experts to each task, resulting in zero BWT loss. This confirms that the router’s embedding clusters (Figure~\ref{fig:router_embeddings}(a)) translate directly into correct expert assignment.
\item \textbf{Constructive Clustering ($P=6$):} Semantically related tasks (e.g., \textit{NumGLUE-cm/ds}) are co-routed to Expert~5, while other tasks retain dedicated experts. Because these merged tasks share similar gradient directions, parameter sharing is beneficial rather than destructive. The BWT curve remains relatively stable with a slight dip at Step~7 when the merge occurs.
\item \textbf{Destructive Overload ($P=4$):} A ``super-expert" absorbs disparate tasks (QA, Stance, Math), leading to conflicting updates and a sharp BWT drop to $-0.19$.  Notably, the router still identifies \textit{Py150} (code) as maximally distinct, preserving its isolation in a separate expert even under severe constraints.
\item \textbf{System Collapse ($P=2$):} All tasks merge into one expert and the BWT curve declines to $-0.25$ as catastrophic forgetting becomes unavoidable. This represents the lower bound of the capacity spectrum and illustrates the fundamental trade-off between memory efficiency and interference prevention.
\end{itemize}

\subsubsection{Pool capacity constraints in the online scenario}
\label{sec:online_pool}
The batched analysis reveals a clear capacity-stability trade-off which we investigate here for the online scenario. We evaluate CaLLM-O over 500-chunk streams with varying pool sizes on TRACE ($P \in \{8, 6, 4\}$) and CITB ($P \in \{19, 10, 3\}$), against SeqFT ($P{=}1$), revealing a non-monotonic relation between performance and capacity. 

\paragraph{Stability over capacity.} On TRACE, $P{=}4$ achieves the 
highest final OMA ($0.42$) and the best BWT, 
outperforming both $P{=}8$ ($0.39$) and $P{=}6$ ($0.38$) (see 
Appendix~\ref{app:online_pool_metrics} for full plots). Routing 
stability---the consistency of task-to-expert assignment---outweighs 
raw capacity (see Appendix ~\ref{app:confusion_matrices} Fig.~\ref{fig:confusion_matrices}). At $P{=}4$, the 
router converges on four semantically coherent groups: (1)~C-STANCE; 
(2)~quantitative tasks (NumGLUE-ds, NumGLUE-cm, Py150); (3)~FOMC; 
(4)~long-form understanding (ScienceQA, MeetingBank, 20Minuten). 
Interleaved chunks within each group provide implicit replay via 
alternating gradient signals, fostering constructive interference. 
Conversely, $P{=}6$ suffers from task displacement: 20Minuten 
is scattered across multiple experts, preventing specialization. 
$P{=}8$ achieves perfect isolation but forgoes the co-assignment 
regularization that benefits the streaming regime.

\paragraph{Similarity and forgetting.} In more homogeneous streams 
like CITB-19, performance is less sensitive to capacity; since similar tasks tolerate compression well, though smaller pools still cause deeper forgetting spikes. Task similarity ultimately determines how sensitive a stream is to pool size, diverse streams need larger, more carefully tuned pools.


\section{Conclusion}
\label{conclusions}

We presented CaLLM, a modular continual learning method for LLMs that pairs a bounded pool of lightweight adapters with a non-parametric, prototype-based router operating entirely in the frozen backbone's embedding space, requiring no gradient-based training or task labels at inference. Incoming data are matched to the nearest expert via cosine distance, with new experts spawned or repurposed automatically when the data is sufficiently novel. Our experiments on TRACE-8 and CITB-19 demonstrate that CaLLM is effective across both batched and online CL settings, where it reliably discovers distributional boundaries from input features alone, and translates this geometric separation into accurate, task-agnostic expert assignments.  Although CITB-19 shows that the router can separate fine-grained tasks that share a language and format, we do not test whether it can keep tasks together when they overlap conceptually but differ lexically (e.g. the same task across languages), which we leave to future evaluation.
Notably, we observe that limiting the adapter pool size can boost performance by forming semantically coherent groups and using interleaved gradients as implicit replay, without requiring memory buffers.

Future directions include integrating online AutoML to adapt currently fixed hyperparameters as the stream unfolds, explicitly merging related adapters to consolidate shared knowledge, and extending prototype-based routing to multi-modal streams spanning vision, language, and other modalities simultaneously.

\bibliographystyle{collas2026_conference}
\bibliography{callm_refs}



\section*{Code Availability}
The code for CaLLM is publicly available at
\url{https://github.com/Bilgecelik/CaLLM-code}.

\section*{Acknowledgments}
Funded by the European Union. Views and opinions expressed are however those
of the author(s) only and do not necessarily reflect those of the European
Union or the European Commission. Neither the European Union nor the European
Commission can be held responsible for them. This work was supported by the
European Union’s Horizon Europe research and innovation programme under grant
agreement No 101214398 (ELLIOT).

\appendix
\clearpage
\section{Appendix}

\subsection{Backbone selection}
\label{app:backbone}

Table~\ref{tab:inference-only} compares zero-shot performance across three candidate backbones -- Llama-3.1-8B-Instruct, gemma-3-12b-it, and Qwen3-8B-- on the TRACE-8 benchmark. While all models struggle on the NumGLUE (math tasks), confirming the necessity of domain-specific adaptation, gemma-3-12b-it consistently exhibits the strongest generalist prior. It achieves the highest zero-shot accuracy across all eight tasks, with marked advantages on C-STANCE ($0.62$ vs.\ $0.20$), FOMC ($0.64$ vs.\ $0.14$), and Py150 ($14.91$ vs.\ $6.42$). Selecting gemma-3-12b-it as our primary backbone ensures that our results evaluate the efficacy of the CaLLM framework rather than compensating for a weak base-model initialization.

\begin{table}[h]
  \centering
  \caption{\textbf{Inference-only (Zero-Shot) Performance on TRACE-8.} gemma-3-12b-it achieves the highest score on all 8 tasks. All models show low performance on math, motivating the need for continual adaptation.}
  \label{tab:inference-only}
  \small
  \begin{tabular}{lcccc}
    \toprule
    \textbf{Task} & \textbf{Metric} & \textbf{Llama-3.1-8B-Instruct} & \textbf{gemma-3-12b-it} & \textbf{Qwen3-8B} \\
    \midrule
    C-STANCE    & acc (0--1)      & 0.20 & \textbf{0.62} & 0.20 \\
    FOMC        & acc (0--1)      & 0.14 & \textbf{0.64} & 0.04 \\
    MeetingBank & ROUGE-L (0--1)  & 0.05 & \textbf{0.21} & 0.04 \\
    Py150       & EDIM (0--100)   & 6.42 & \textbf{14.91} & 0.00 \\
    ScienceQA   & acc (0--1)      & 0.09 & \textbf{0.35} & 0.07 \\
    NumGLUE-cm  & acc (0--1)      & 0.01 & \textbf{0.05} & 0.00 \\
    NumGLUE-ds  & acc (0--1)      & 0.02 & \textbf{0.07} & 0.00 \\
    20Minuten   & SARI (0--100)   & 36.2 & \textbf{39.22} & 37.0 \\
    \bottomrule
  \end{tabular}
\end{table}

\subsection{Implementation details and hyperparameters}
\label{app:implementation}

As detailed in Appendix~\ref{app:backbone}, all experiments use gemma-3-12b-it (4-bit quantized) as the primary backbone. We maintain this fixed model for all comparisons, though CaLLM is backbone-agnostic and compatible with any Transformer-based LLM. The router utilizes the same frozen backbone as a feature extractor, where the last hidden states are mean-pooled over valid tokens to obtain a sample embedding and then averaged across the input batch to produce a single representation for routing.

LoRA adapters are trained using the HuggingFace SFTTrainer with an 8-bit AdamW optimizer, gradient clipping at norm~$1.0$, and bfloat16 mixed precision without sequence packing. Additional training constants include a maximum sequence length of $1024$, temperature of $1.0$, LoRA dropout of $0.05$, and a warmup ratio of $0.05$. In the online setting, we use $max\_chunks=500$ and $chunk\_size=40$, with chunks generated via random resampling. All input follows the Alpaca prompt style detailed in \S\ref{sec:setup}, and the router uses the adaptive threshold defined in Eq.~\ref{eq:adaptive_threshold}. 

Batched experiments are reported as the mean and standard deviation across three random seeds with optimal configurations selected based on final mean validation performance. Online experiments are instead performed as a single run due to the significant computational cost of $500$-chunk streams. We rely on the OMA metric, which averages over $500$ evaluation steps, to provide inherent statistical stability. Table~\ref{tab:hyperparams} summarizes the specific hyperparameters selected via grid search over learning rate $\in\{10^{-4}, 5\times 10^{-4}, 10^{-3}\}$, LoRA rank $\in\{4, 8, 16\}$, and epochs $\in\{3, 5\}$. To mitigate the substantial computational overhead of tuning long-horizon streams, hyperparameter optimization for the online setting was restricted to a truncated proxy window of 200 evaluation steps under a standard pre-update (test-than-train) evaluation protocol.
All experiments were conducted on a single NVIDIA RTX A6000 GPU with 40GB of VRAM using \texttt{Unsloth} \citep{unsloth} and the \texttt{trl}, \texttt{peft} and \texttt{transformers} libraries from HuggingFace.

\begin{table}[h]
  \centering
  \caption{\textbf{CaLLM hyperparameters per scenario.} All runs use the HuggingFace SFTTrainer with 8-bit AdamW, gradient clipping at norm~1.0, and bfloat16 mixed precision. LR = learning rate, $r$ / $\alpha$ = LoRA rank / scaling factor, Batch = batch size, GAS = gradient accumulation steps, LR Sched. = learning rate schedule.}
  \label{tab:hyperparams}
  \small
  \begin{tabular}{lcccccccc}
    \toprule
    \textbf{Setting} & \textbf{LR} & \textbf{$r$} & \textbf{$\alpha$} & \textbf{Epochs} & \textbf{Batch} & \textbf{GAS} & \textbf{LR Sched.} \\
    \midrule
    CaLLM-B, TRACE  & $1\!\times\!10^{-4}$ & 16 & 32 & 3 & 16 & 4 & cosine  \\
    CaLLM-B, CITB   & $1\!\times\!10^{-3}$ & 8  & 16 & 5 & 25 & 4 & cosine  \\
    \midrule
    CaLLM-O, TRACE  & $3\!\times\!10^{-3}$ & 4  & 8  & 1 & 10 & 2 & linear \\
    CaLLM-O, CITB   & $1\!\times\!10^{-3}$ & 8  & 8  & 1 & 20 & 1 & cosine\\
    \bottomrule
  \end{tabular}
\end{table}

\subsection{Details about the baselines}
\label{app:baselines}

To evaluate CaLLM, we compare it against several representative strategies from the CL literature, adapted for the PEFT regime. To ensure a fair and rigorous comparison, all baselines share the same core architectural parameters as CaLLM and the same LoRA configuration (rank, alpha, and dropout). We then utilize the optimal hyperparameters (such as regularization strength or buffer size) specific to each method as reported in their respective original papers, unless stated otherwise.

To understand the potential impact of this design choice on the fairness of baseline comparisons, we conducted an ablation study of hyperparameter search on selected continual learning baselines, namely ER and EWC. The search follows the CaLLM search space over learning rate and number of training epochs, using a grid search over learning rate \(\in \{10^{-4}, 5 \times 10^{-4}, 10^{-3}\}\) and epochs \(\in \{3, 5\}\). The ablation focuses on the TRACE benchmark, due to its higher variation across tasks and smaller number of tasks. Results of this search are presented in Table~\ref{tab:baseline_hp_ablation}, including the original hyperparameter values and performance results. As shown in the table, the search selects a hyperparameter combination different from the originally chosen values for each baseline (where the per-task performance does not always improve compared with the original setting). However, the magnitude of the performance difference is small, and may also be explained by the inherent randomness of LLM generations. Importantly, this does not change the ordering of baseline--method comparisons for any task in the analysis, supporting the original design choice.

\begin{table}[t]
\centering
\caption{Baseline hyperparameter ablation on TRACE-8 (Batched). Original rows use the configuration from the main comparison; tuned rows are selected from the same learning-rate/epoch grid used for CaLLM while keeping LoRA rank, scaling factor, and dropout fixed.}
\label{tab:baseline_hp_ablation}
\resizebox{\textwidth}{!}{%
\begin{tabular}{llcccccccccc}
\toprule
Method & Config. & LR & Epochs & C-STANCE & FOMC & MeetingBank & Py150 & ScienceQA & NumGLUE-cm & NumGLUE-ds & 20Minuten \\
\midrule
ER  & Original & $1\times10^{-4}$ & 3 & 0.04 & 0.14 & 0.10 & 12.98 & 0.19 & 0.06 & 0.13 & 39.63 \\
ER  & Tuned    & $5\times10^{-4}$ & 5 & 0.11 & 0.19 & 0.12 & 13.04 & 0.22 & 0.09 & 0.15 & 39.58 \\
EWC & Original & $1\times10^{-4}$ & 3 & 0.28 & 0.31 & 0.10 & 10.55 & 0.61 & 0.12 & 0.15 & 40.06 \\
EWC & Tuned    & $5\times10^{-4}$ & 3 & 0.17 & 0.35 & 0.09 & 10.61 & 0.55 & 0.21 & 0.11 & 39.01 \\
\bottomrule
\end{tabular}%
}
\end{table}

\begin{itemize}
    \item \textbf{Inference only} is a simple baseline where the base model (e.g. gemma-3-12b-it) is used during inference in a continual/online way, every time a new task/chunk is presented to the model.
    \item \textbf{Sequential Adapter Fine-Tuning (Seq-FT)}: A single LoRA adapter is fine-tuned sequentially on the task stream. This represents the lower bound for stability, as it is highly susceptible to catastrophic forgetting. 
    \item \textbf{Elastic Weight Consolidation (EWC) \citep{Kirkpatrick2017EWC}}: A regularization-based approach that penalizes changes to parameters that were important for previous tasks. Importance is estimated via the diagonal of the Fisher Information Matrix. We apply EWC to the LoRA weights specifically.
    \item \textbf{Experience Replay (ER) \citep{Rolnick2019ER}:} A memory buffer stores a subset of samples from previous tasks. During the adaptation to a new task, each training batch is augmented with replayed samples from the buffer to preserve past knowledge. In our experiments, we use a fixed buffer size of $1000$ samples.
    \item \textbf{Orthogonal LoRA (OLoRA) \citep{Wang2023OLoRA}:} A subspace isolation method that constrains the LoRA weight updates to be orthogonal to the pre-trained weights or previous task directions, aiming to prevent gradient-based interference. To ensure compatibility with the Gemma backbone and avoid out-of-memory (OOM) errors, we modified the original OLoRA implementation as follows: (i)~reduced the evaluation batch size to $4$, instead of using the batch size, and (ii)~offloaded the orthogonal loss calculation to the CPU. For the TRACE-8 online stream, we utilized $batch\_size=8$, $r=4$, $\alpha=8$. For CITB, we used $batch\_size=2$, $r=4$, $\alpha=8$, and gradient accumulation step of $8$. 
    \item \textbf{K-D Tree of Low-Rank Adapters (TreeLoRA)~\citep{qian2025treelora}:} Introduces layer-wise adapters to construct hierarchical tree based on the criteria of gradient similarity, exploring task relationships via lower-confidence-bound bandits.
\end{itemize}

\subsection{gemma-3-12b-it training data and CITB overlap}
\label{app:citb_overlap}
Gemma 3 models are well-suited for text-generation tasks such as question answering, summarization, and reasoning. The 12B model used in our work was trained on 12T tokens consisting of web documents (in over 140 languages), code, mathematics, and images. The pre-training data is not released openly and there is limited information about the data sources. Additionally, the IT variant has been instruction-tuned using knowledge distillation from larger teacher models (likely Gemini models).

CITB is built on top of the Super-NaturalInstructions dataset \citep{wang2022supernaturalinstructions}. In this work, we focus on the InstrDialog task stream, which consists of 19 dialogue-related tasks spanning three categories: dialogue state tracking (4 tasks), dialogue generation (11 tasks), and intent identification (4 tasks).
All tasks are evaluated using the ROUGE-L metric.

While there is no clear information about the data used by Gemma~3 for pre-training or fine-tuning, it is plausible that the  Super-NaturalInstructions dataset (on which CITB is based) was included in the training data of either Gemma~3 or the Gemini teacher models. This would explain why LoRA fine-tuning on CITB does not improve over the inference-only baseline. Despite this, all other baselines suffer from sequential fine-tuning, suggesting that CaLLM maintains the original performance even when the data has already been seen.

To investigate potential data contamination, we applied the method of ~\citet{Shi2023Detecting}, which computes per-token probabilities under the hypothesis that ``seen" examples have fewer low-probability outliers with respect to ``unseen" words. We compared 500 CITB samples against 309 ``old" Wikipedia pages (likely within Gemma~3's training data) and 97 Wikipedia pages created in the following 60 days (unlikely to be in the training data).
The results are as follows:
\begin{itemize}
    \item Member mean:      0.005722
    \item Non-member mean:  0.003475
    \item CITB mean:        0.002848
\end{itemize}
From these results we can conclude that CITB has an average probability closer to the non-member samples than the (assumed) member samples, suggesting that CITB is likely not in the training data of Gemma~3.

This appears to contradict our earlier hypothesis that Gemma~3 has already seen CITB data. Indeed, Figure~\ref{fig:batched_baseline_comparison}b shows that fine-tuning with CITB data does not improve over the inference-only baseline. However, it is important to note that the method of \citet{Shi2023Detecting} is only an empirical heuristic, and the candidate member and non-member sets are based on assumptions about Wikipedia page creation dates rather than ground-truth training data membership.

\subsection{Synthetic data generation for CITB}\label{app:synthetic_data}

The original CITB benchmark, defined as InstrDialog in the original paper \citep{CITB2023}, consists of 19 tasks derived from Super-NaturalInstructions, detailed in Table~\ref{tab:citb_tasks}.

\begin{table}[h]
\centering
\caption{The 19 dialogue tasks in the CITB InstrDialog stream.}
\label{tab:citb_tasks}
\begin{small}
\begin{tabular}{llc}
\toprule
\textbf{Category} & \textbf{Task Type} & \textbf{Count} \\
\midrule
Dialogue State Tracking & Slot Filling, State Update & 4 \\
Dialogue Generation & Response Gen, Clarification & 11 \\
Intent Identification & Classification, Intent Detection & 4 \\
\midrule
\textbf{Total} & & \textbf{19} \\
\bottomrule
\end{tabular}
\end{small}
\end{table}

Each task contains only $500$ training instances, which is insufficient for effectively training LoRA adapters and creates an imbalance when compared to the TRACE-8 benchmark ($5000$ instances per task). To ensure a fair comparison and improve adaptation stability, we generated synthetic training data for each CITB task. Inspired by self-synthesized rehearsal strategies \citep{Huang2024SSR}, we utilized the frozen gemma-3-12b-it model as a data generator. We employed a few-shot prompting strategy, providing the model with two randomly sampled real instances in-context and tasking it to generate a novel example following the same Alpaca-style format. By iterating this process with randomized in-context examples, we produced an initial pool of $6000$ synthetic instances per task.To ensure data quality and distributional consistency, we applied a two-stage refinement pipeline:
\begin{enumerate}
    \item \textbf{Deduplication:} We removed all exact string matches and near-duplicates to ensure diversity within the synthetic pool.
    \item \textbf{Centroid-based filtering:} To retain only high-quality, representative samples, we converted the synthetic instances into embeddings using the frozen base model. We then applied K-means clustering to the embeddings of each task. Finally, we calculated the Euclidean distance between each instance and its respective cluster centroid, selecting the $3000$ instances closest to the centroid for the final training set.
\end{enumerate}

This procedure resulted in a refined CITB dataset containing $3000$ examples per task. This refined version maintains the semantic integrity of the original tasks while providing sufficient signal for parameter-efficient fine-tuning, and it serves as the standard CITB data stream used for all experiments and baseline comparisons presented in this paper.

\FloatBarrier
\subsection{Automatic Metric Selection}
\label{app:auto_metric}

When ground-truth metric labels are unavailable (i.e., in label-free stream configurations), CaLLM dispatches an automatic metric selection procedure at the start of each task's evaluation phase. This adds only a negligible time overhead from a single short inference call and no memory overhead, as it reuses the already-loaded backbone. The procedure works as follows.

\paragraph{Model.} The same base LLM used for task inference (gemma-3-12b-it, 4-bit quantized) serves as the zero-shot classifier for metric selection. This avoid the overhead of loading an auxiliary model and the system simply executes a single generation call before the model training and evaluation on each task. 

\paragraph{Meta-prompt template.} A single representative prompt is sampled from the task's test split, truncated to 700 characters to manage context, and embedded into a structured meta-prompt. The prompt instructs the LLM to categorize the task based on its linguistic and structural features (e.g., presence of code-specific keywords, multiple-choice indicators, or summarization requests). The model is constrained to a closed set of responses: \texttt{ACCURACY}, \texttt{ROUGE}, \texttt{SARI}, or \texttt{EDIM}.

\begin{quote}
\small
\texttt{Analyze the following task prompt and select the best evaluation metric.}\\[2pt]
\texttt{RULES:}\\
\texttt{1. Code completion tasks (contains `code', `python', `complete the code', etc.) $\rightarrow$ Use EDIM}\\
\texttt{2. Summarization tasks (contains `summarize', `summary', `write a summary', etc.) $\rightarrow$ Use ROUGE}\\
\texttt{3. Multiple choice questions (contains `A)', `B)', `C)', `choose', etc.) $\rightarrow$ Use ACCURACY}\\
\texttt{4. Text simplification tasks (contains `simplify', `simplified version', etc.) $\rightarrow$ Use SARI}\\
\texttt{5. Default for all other tasks $\rightarrow$ Use ACCURACY}\\[2pt]
\texttt{Task prompt: \{prompt\}}\\[2pt]
\texttt{Respond with ONLY the metric name. Choose one: ACCURACY, ROUGE, SARI, or EDIM}\\
\texttt{\#\#\# Response:}
\end{quote}

\paragraph{Parsing and fallback.} To ensure robustness, the model's output is parsed for the first occurrence of a valid metric keyword. If the model fails to produce a recognized metric name within its response, the system defaults to \texttt{ACCURACY} as a conservative baseline. Once a metric is selected, it is cached for the duration of the task's residency in the stream to prevent redundant computation.

\clearpage
\section{Algorithms and Efficiency}

\subsection{Algorithms (pseudocode)}
\label{app:algorithms}
\begin{algorithmbox}{CaLLM-Batched}
\label{alg:batched}
\textbf{Input:} Task sequence $\mathcal{D}_1, \ldots, \mathcal{D}_T$;\; threshold $\theta$ (fixed or adaptive, Eq.~\ref{eq:adaptive_threshold});\; pool capacity $P$\\
\textbf{Output:} Performance table $\mathbf{M}$\\[4pt]
Initialize pool $\mathcal{P} = \varnothing$\\[2pt]
\textbf{for} $t = 1$ \textbf{to} $T$ \textbf{do}\\[2pt]
\quad \textit{// Route (training mode: prototype updates enabled)}\\
\quad $\mathbf{e} \leftarrow \textsc{Embed}(\text{sample}(\mathcal{D}_t^{\text{train}}))$\\
\quad $i^*,\, d^* \leftarrow \arg\min_{i : \ell_i \in \mathcal{P}}\,d_i$
\quad (Eq.~\ref{eq:cosine_dist};\; $d^* = \infty$ if $\mathcal{P} = \varnothing$)\\
\quad \textbf{if} $d^* \leq \theta$ \textbf{then}\\
\quad\quad $a_t \leftarrow \ell_{i^*}$;\;
           $\mathbf{p}_{i^*} \leftarrow \tfrac{1}{2}(\mathbf{p}_{i^*} + \mathbf{e})$;\;
           $c_{i^*} \leftarrow c_{i^*} + 1$
\quad \textit{// reuse}\\
\quad \textbf{else if} $|\mathcal{P}| < P$ \textbf{then}\\
\quad\quad $a_t \leftarrow \textsc{NewAdapter}()$;\;
           $\mathcal{P} \leftarrow \mathcal{P} \cup \{(a_t,\, \mathbf{e},\, 1)\}$
\quad \textit{// create}\\
\quad \textbf{else}\\
\quad\quad $i_{\min} \leftarrow \arg\min_{i} c_i$;\;
           $a_t \leftarrow \ell_{i_{\min}}$;\;
           $\mathbf{p}_{i_{\min}} \leftarrow \mathbf{e}$;\;
           $c_{i_{\min}} \leftarrow 1$
\quad \textit{// repurpose}\\
\quad \textbf{end if}\\[2pt]
\quad \textit{// Train (overwrites adapter weights in the repurposing case)}\\
\quad $\textsc{SFT}(\mathcal{D}_t^{\text{train}},\, \mathcal{D}_t^{\text{eval}},\, \text{adapter}{=}a_t)$\\[2pt]
\quad \textit{// Evaluate current task}\\
\quad $\textsc{LoadAdapter}(a_t)$;\;
      $\mathbf{M}[t, t] \leftarrow \textsc{Evaluate}(\mathcal{D}_t^{\text{test}})$\\[2pt]
\quad \textit{// Backward transfer on all previous tasks (read-only routing)}\\
\quad \textbf{if} $t > 1$ \textbf{then}\\
\quad\quad \textbf{for} $j = 1$ \textbf{to} $t-1$ \textbf{do}\\
\quad\quad\quad $\mathbf{e}_j \leftarrow \textsc{Embed}(\text{sample}(\mathcal{D}_j^{\text{test}}))$\\
\quad\quad\quad $a_j \leftarrow \ell_{\arg\min_{i} d_i(\mathbf{e}_j)}$
               \quad \textit{// read-only: no prototype updates}\\
\quad\quad\quad $\textsc{LoadAdapter}(a_j)$;\;
               $\mathbf{M}[j, t] \leftarrow \textsc{Evaluate}(\mathcal{D}_j^{\text{test}})$\\
\quad\quad \textbf{end for}\\
\quad\quad $\mathbf{M}[\text{BWT}, t] \leftarrow
           \frac{1}{t-1}\sum_{j=1}^{t-1}
           \bigl(\mathbf{M}[j,t] - \mathbf{M}[j,j]\bigr)\,/\,s_j$\\
\quad \textbf{else}\; $\mathbf{M}[\text{BWT}, t] \leftarrow 0$\\
\quad \textbf{end if}\\[2pt]
\quad $\mathbf{M}[\text{Avg}, t] \leftarrow
      \frac{1}{t}\sum_{i=1}^{t} \mathbf{M}[i,t]\,/\,s_i$\\
\textbf{end for}
\end{algorithmbox}

\begin{algorithmbox}{CaLLM-Online}
\label{alg:online}

\textbf{Input:} Chunk stream $(\tau_k, \mathcal{D}_k)$ for $k = 1, \ldots, N$;\;
               threshold $\theta$ (fixed or adaptive);\; pool capacity $P$\\
\textbf{Output:} Performance table $\mathbf{M}$ with per-step Forgetting, BWT, and OMA\\[4pt]
Initialize: pool $\mathcal{P} = \varnothing$,\; $\mathrm{last} = \{\}$,\;
            $S_{\text{run}} = 0$,\; $c = 0$\\[2pt]
\textbf{for each} step $k = 1, \ldots, N$, receiving $(\tau, \mathcal{D})$ \textbf{do}\\[2pt]
\quad \textit{// Route (single pass: prototype updates enabled)}\\
\quad $\mathbf{e} \leftarrow \textsc{Embed}(\text{sample}(\mathcal{D}))$\\
\quad $i^*,\, d^* \leftarrow \arg\min_{i : \ell_i \in \mathcal{P}}\,d_i$
\quad ($d^* = \infty$ if $\mathcal{P} = \varnothing$)\\
\quad \textbf{if} $d^* \leq \theta$ \textbf{then}\\
\quad\quad $a \leftarrow \ell_{i^*}$;\;
           $\mathbf{p}_{i^*} \leftarrow \tfrac{1}{2}(\mathbf{p}_{i^*} + \mathbf{e})$;\;
           $c_{i^*} \leftarrow c_{i^*} + 1$
\quad \textit{// reuse}\\
\quad \textbf{else if} $|\mathcal{P}| < P$ \textbf{then}\\
\quad\quad $a \leftarrow \textsc{NewAdapter}()$;\;
           $\mathcal{P} \leftarrow \mathcal{P} \cup \{(a,\, \mathbf{e},\, 1)\}$
\quad \textit{// create}\\
\quad \textbf{else}\\
\quad\quad $i_{\min} \leftarrow \arg\min_{i} c_i$;\;
           $a \leftarrow \ell_{i_{\min}}$;\;
           $\mathbf{p}_{i_{\min}} \leftarrow \mathbf{e}$;\;
           $c_{i_{\min}} \leftarrow 1$
\quad \textit{// repurpose}\\
\quad \textbf{end if}\\[2pt]
\quad \textit{// Pre-update evaluation (test-then-train)}\\
\quad $\textsc{LoadAdapter}(a)$;\;
      $\mathrm{pre} \leftarrow \textsc{Evaluate}(\mathcal{D})$;\;
      $\mathbf{M}[\tau\text{(pre)}, k] \leftarrow \mathrm{pre}$\\[2pt]
\quad \textit{// BWT (defined only when task has a prior observation)}\\
\quad \textbf{if} $\tau \in \mathrm{last}$ \textbf{then}\\
\quad\quad $\mathbf{M}[\text{BWT}, k] \leftarrow
           (\mathrm{pre} - \mathrm{last}[\tau])\,/\,s_\tau$\\
\quad \textbf{end if}\\[2pt]
\quad $\mathrm{last}[\tau] \leftarrow \mathrm{pre}$\\[2pt]
\quad $S_{\text{run}} \leftarrow S_{\text{run}} + \mathrm{pre}\,/\,s_\tau$;\;
      $c \leftarrow c + 1$;\;
      $\mathbf{M}[\text{OMA}, k] \leftarrow S_{\text{run}}\,/\,c$\\[2pt]
\quad \textit{// Train on same adapter, same chunk}\\
\quad $\textsc{SFT}(\mathcal{D},\, \text{adapter}{=}a)$\\
\textbf{end for}
\end{algorithmbox}

\FloatBarrier
\subsection{Computational and Memory Efficiency}
\label{app:compute}

A practical CL system must not only preserve performance but also remain efficient as the expert pool grows. Table~\ref{tab:compute} reports training FLOPs, throughput, and peak GPU memory for CaLLM at varying pool sizes alongside baselines, measured on the full TRACE-8 batched pipeline (gemma-3-12b-it, single NVIDIA RTX A6000 48\,GB GPU, CUDA 12.4).

\begin{table}[h]
  \centering
  \caption{\textbf{Compute and Memory Overhead on TRACE-8 (Batched).} CaLLM consumes the same training FLOPs as single-adapter Seq-FT regardless of pool size, with constant GPU memory. ER doubles wall-clock time due to replay, and oLoRA requires $2.1\times$ more GPU memory for orthogonal projections. $^\dagger$EWC, ER, and oLoRA use custom training loops that do not report \texttt{total\_flos}; EWC's FLOPs are comparable to CaLLM (same data/rank), ER's are higher due to replay augmentation, and oLoRA's are lower due to smaller rank ($r{=}4$).}
  \label{tab:compute}
  \small
  \begin{tabular}{lcccccc}
    \toprule
    \textbf{Method} & \textbf{Pool $P$} & \textbf{Train TFLOPs} & \textbf{Samples/s} & \textbf{Peak GPU (GB)} & \textbf{Wall (h)} & \textbf{\# Adapters} \\
    \midrule
    CaLLM-B & 2 & 7,339 & 3.81 & 12.15 & 60.0 & 2 \\
    CaLLM-B & 4 & 7,339 & 3.81 & 12.15 & 59.3 & 4 \\
    CaLLM-B & 6 & 7,339 & 3.67 & 12.15 & 59.7 & 6 \\
    CaLLM-B & 8 & 7,339 & 3.87 & 12.15 & 59.5 & 8 \\
    \midrule
    Seq-FT     & 1 & 7,339 & 3.53 & 12.15 & 68.2 & 1 \\
    EWC$^\dagger$  & 1 & ---   & ---  & 12.48 & 76.9 & 1 \\
    ER$^\dagger$   & 1 & ---   & ---  & 12.10 & 125.8 & 1 \\
    oLoRA$^\dagger$ & 8 & ---  & ---  & 25.20 & 75.1 & 8 \\
    \bottomrule
  \end{tabular}
\end{table}

\paragraph{Training compute.}
The central result is that CaLLM's training cost is \emph{independent of pool size}: all four configurations ($P = 2, 4, 6, 8$) consume exactly 7,339\,TFLOPs. This is a direct consequence of the modular architecture---only one adapter is active per forward pass, so the computational graph is identical regardless of how many adapters exist in the pool. Training throughput is also stable across pool sizes (3.67--3.87\,samples/s), confirming that adapter switching introduces negligible latency.

CaLLM and Seq-FT share identical training FLOPs (7,339\,TFLOPs), as expected---both perform the same number of gradient updates on the same data with the same adapter rank. The difference lies entirely in \emph{where} gradients are routed, not in how many are computed. CaLLM achieves slightly higher throughput (3.81--3.87 vs.\ 3.53\,samples/s) and lower wall-clock time (59--60\,h vs.\ 68\,h), likely because isolated adapters encounter simpler optimization landscapes than a shared adapter accumulating all tasks.
\paragraph{Routing overhead.} Unlike the baselines, CaLLM performs an additional forward pass through the frozen backbone for prototype matching (\S~\ref{sec:router}). Because this routing pass is encoder-only, i.e., no decoding, over a short input and reuses the already-loaded backbone, its marginal cost is small relative to the per-task training and inference compute. Using the standard transformer estimate of approximately \(2N\) FLOPs per forward token, where \(N \approx 12 \times 10^{9}\) for \texttt{gemma-3-12b-it}, and noting that each routing call processes \(n \approx 20\) prompts of approximately \(10\) tokens each, one routing call costs \(2 \cdot (12 \times 10^{9}) \cdot 20 \cdot 10 \approx 4.8 \times 10^{12}\) FLOPs, i.e., \(4.8\) TFLOPs. Over the full TRACE-8 batched run, with one routing call per task across \(8\) tasks, the router consumes \(8 \cdot 4.8 \approx 38.4\) TFLOPs, or roughly \((38.4 / 7339) \times 100 \approx 0.5\%\) of the \(7339\) TFLOPs reported above for training. This is consistent with the observation that CaLLM and Seq-FT report identical training TFLOPs in Table~\ref{tab:compute}: the router's contribution falls below the reporting precision. The same logic applies at inference time, where each routing call is amortized over all subsequent evaluation queries within a task, giving per-query overhead \(\ll 0.1\%\), and during the online continual-learning phase, where routing is invoked once per chunk rather than per query. The routing module is, moreover, a tunable component: a smaller encoder or cached prompt embeddings could reduce this already-small overhead further, which we leave to future work.

\paragraph{Baseline comparison.}  Among the baselines that explicitly address forgetting, each pays a distinct overhead. EWC trains on the same data with the same adapter rank, so its per-step FLOPs are expected to be comparable to CaLLM; the additional wall-clock cost (76.9\,h) comes from Fisher information computation after each task. ER augments each training batch with replay buffer samples, which increases the effective data budget and thus is expected to incur higher total training FLOPs---this is reflected in its $2.1\times$ wall-clock overhead (125.8\,h). oLoRA uses a smaller rank ($r{=}4$ vs.\ $r{=}16$), which should reduce per-step FLOPs, but must maintain orthogonal projection matrices across all subspaces simultaneously, making it the most memory-intensive baseline at 25.2\,GB ($2.1\times$ CaLLM).

These results confirm that CaLLM achieves its anti-forgetting benefits at \emph{zero additional training compute} relative to na\"{i}ve sequential fine-tuning, while the baselines that attempt to mitigate forgetting through regularization, replay, or orthogonality constraints each pay a substantial cost in time, memory, or both.

\paragraph{Memory management.}
CaLLM maintains a constant 12.15\,GB GPU memory footprint regardless of pool size, because exactly one adapter is loaded on GPU at any time. Inactive adapters are offloaded to CPU memory and swapped in on demand, so the GPU overhead does not grow with the number of adapters in the pool. The maximum pool capacity $P_{\max}$ can be set explicitly or computed automatically based on available system memory according to Eq.~\ref{eq:pool_size}:

\begin{equation}
    P = \left\lfloor \frac{M_{\text{avail}}}{S_{\text{adapter}}} \right\rfloor - 1,
    \label{eq:pool_size}
\end{equation}

where $M_{\text{avail}}$ is the available system (CPU) memory and $S_{\text{adapter}}$ is the on-disk size of a single adapter (both in GB); the $-1$ reserves one slot for the adapter currently resident on GPU. In the online regime, peak GPU memory remains flat at 12.01\,GB---virtually identical to the batched setting---confirming that this constant-memory property holds across both training protocols.

\paragraph{Online regime.}
The online streaming configuration uses lighter adapters ($r{=}4$, $\alpha{=}8$) and single-epoch chunk updates, so its total FLOPs are not directly comparable to the batched runs above. A notable characteristic of the online setting is that \emph{evaluation dominates the runtime}: out of 19.3\,h total wall-clock time over 500 chunks, only 3.6\,h (18.4\%) is spent on training while 13.9\,h (72.3\%) is consumed by the per-chunk test-then-train evaluation loop. This suggests that evaluation scheduling---rather than training efficiency---is the primary lever for reducing online deployment cost, and motivates future work on sparse or amortized evaluation strategies.

\FloatBarrier
\section{Extended Quantitative Results}
\subsection{Online Pool Constraints: OMA and BWT Curves}
\label{app:online_pool_metrics}

\begin{figure}[h]
\centering
\begin{subfigure}[b]{0.48\linewidth}
  \includegraphics[width=\linewidth]{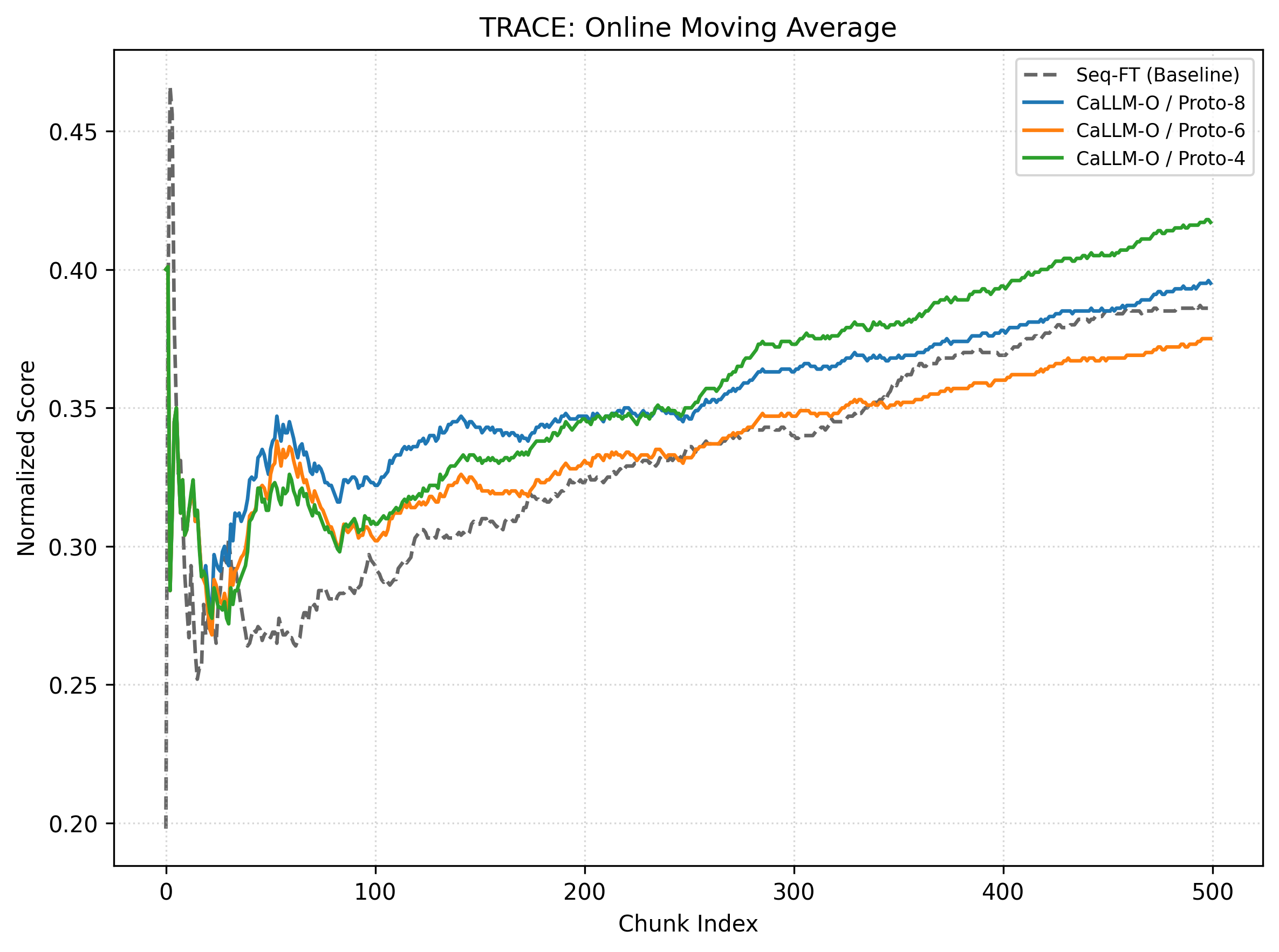}
  \caption{TRACE OMA ($P \in \{4,6,8\}$)}
  \label{fig:online_trace_oma}
\end{subfigure}
\hfill
\begin{subfigure}[b]{0.48\linewidth}
  \includegraphics[width=\linewidth]{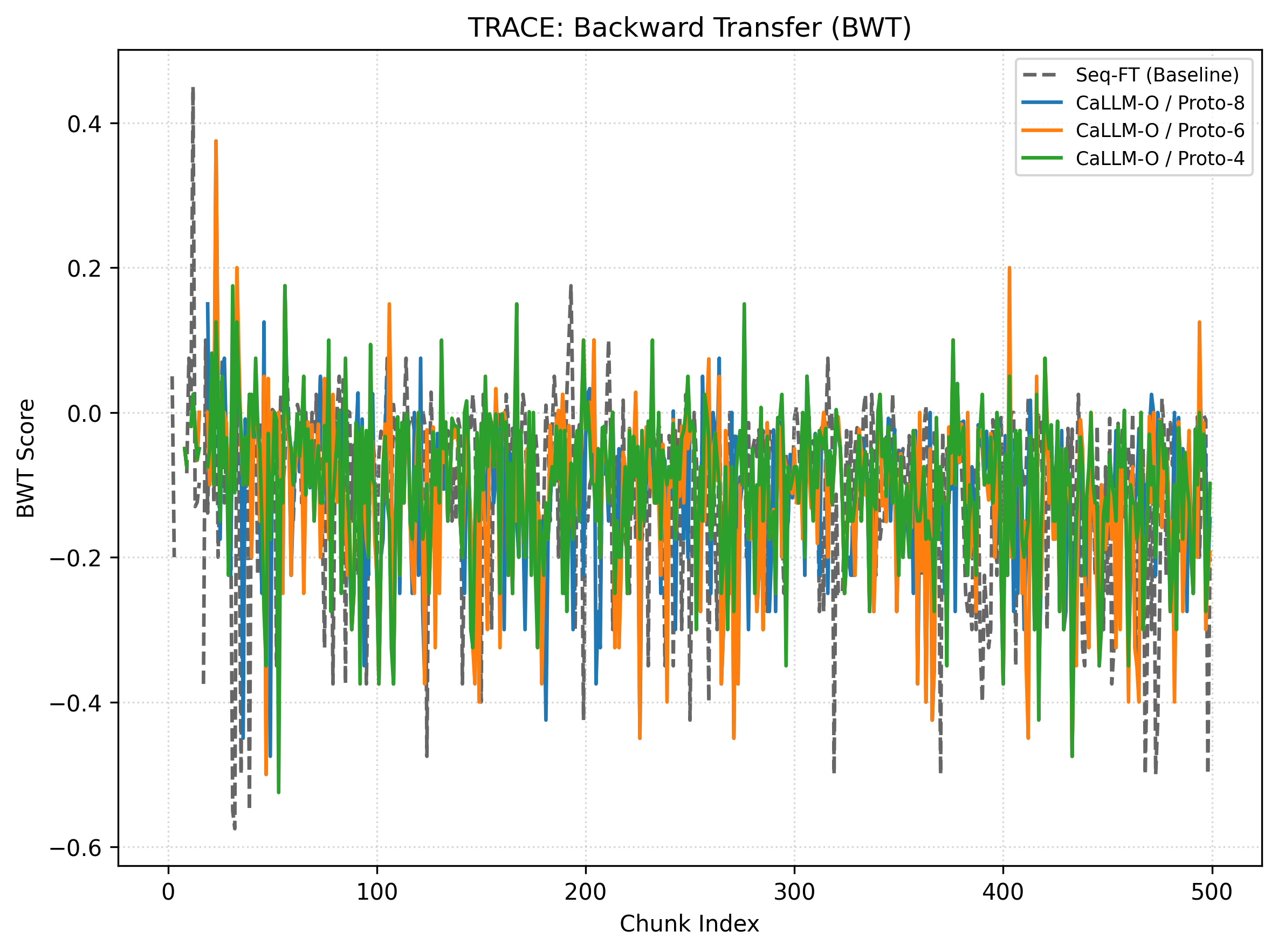}
  \caption{TRACE BWT ($P \in \{4,6,8\}$)}
  \label{fig:online_trace_bwt}
\end{subfigure}
\\[0.4em]
\begin{subfigure}[b]{0.48\linewidth}
  \includegraphics[width=\linewidth]{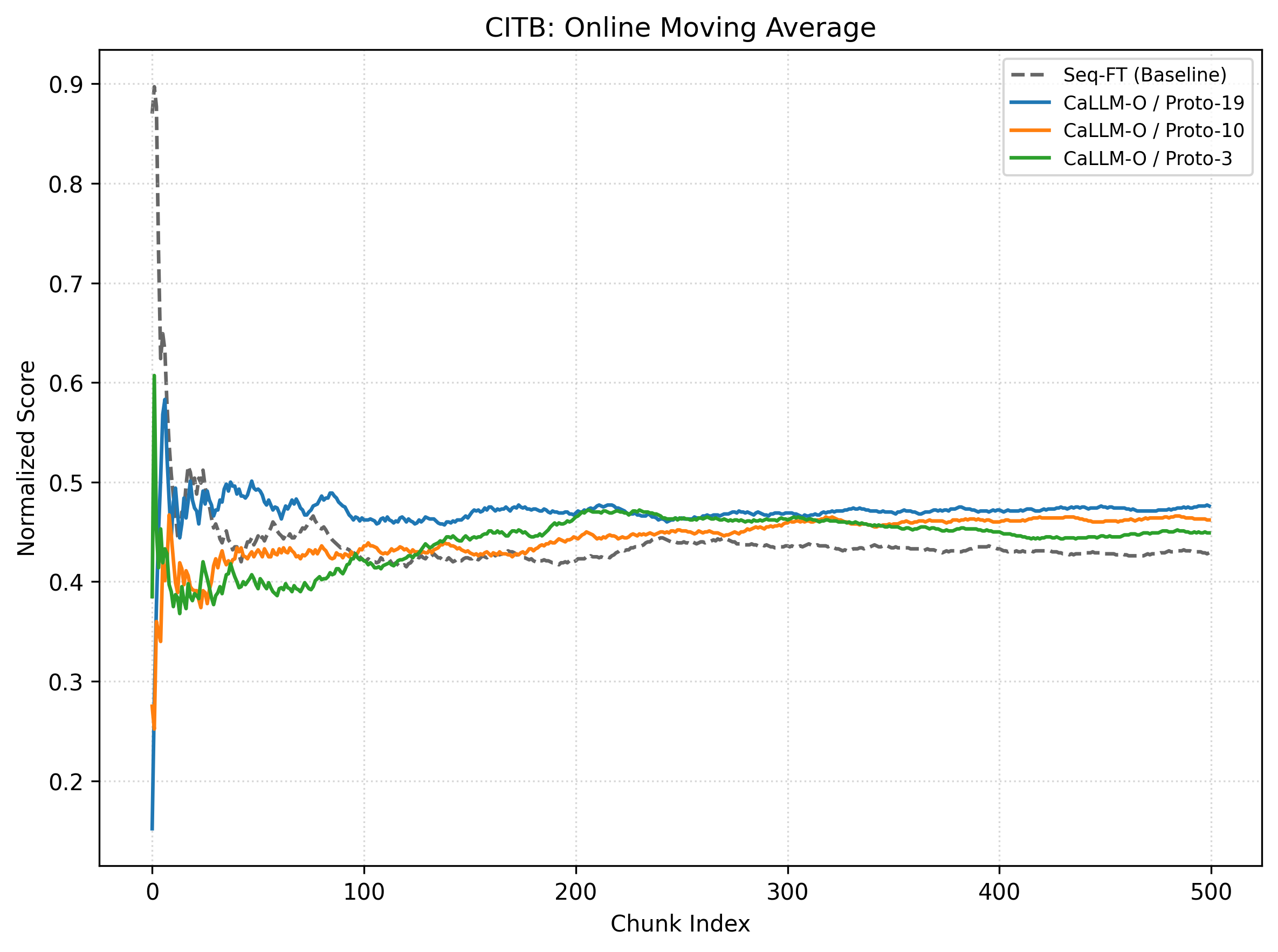}
  \caption{CITB OMA ($P \in \{3,10,19\}$)}
  \label{fig:online_citb_oma}
\end{subfigure}
\hfill
\begin{subfigure}[b]{0.48\linewidth}
  \includegraphics[width=\linewidth]{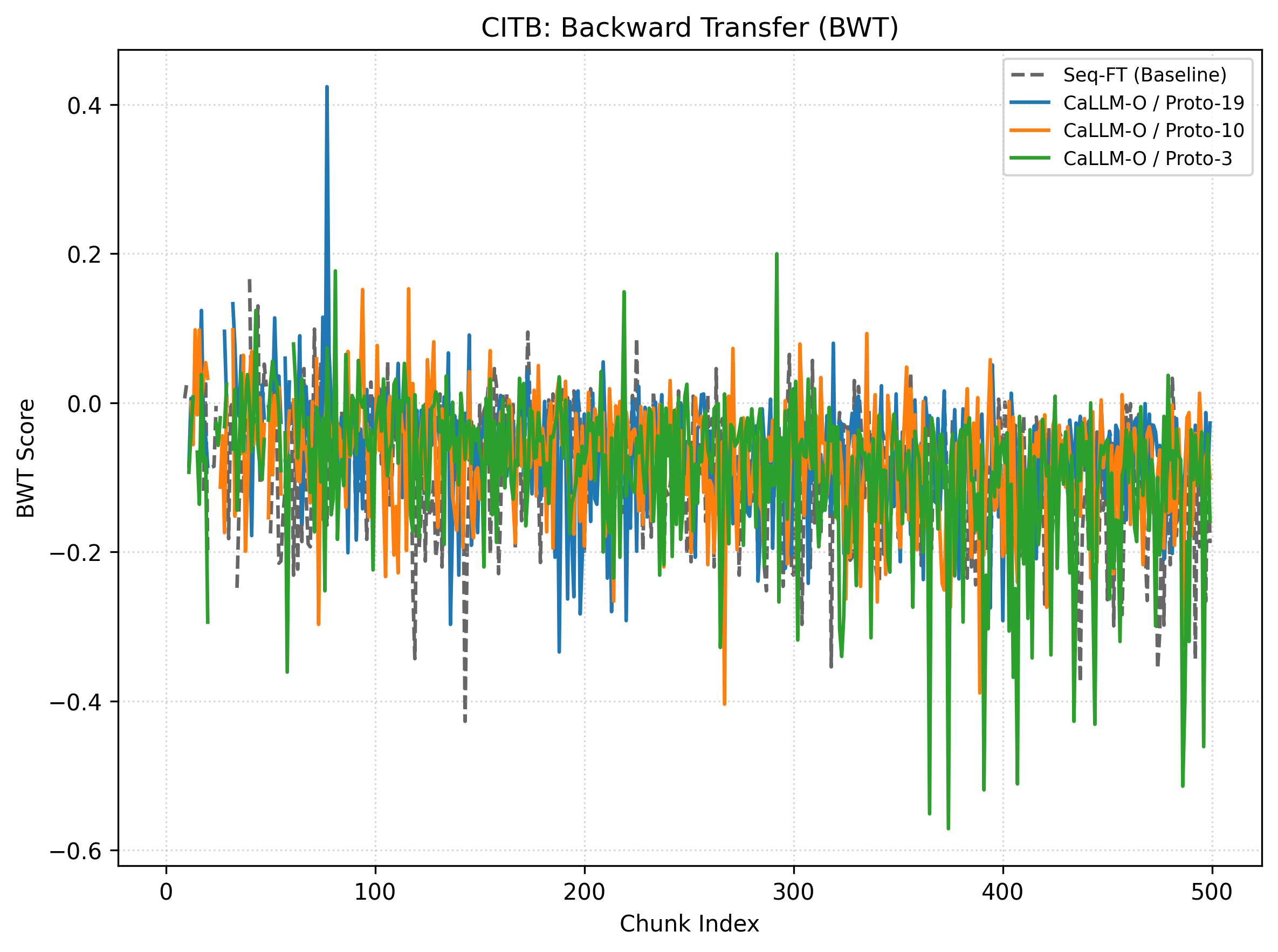}
  \caption{CITB BWT ($P \in \{3,10,19\}$)}
  \label{fig:online_citb_bwt}
\end{subfigure}
\caption{\textbf{Online Pool Constraints (500-chunk stream).} OMA and BWT for CaLLM-O and the SeqFT baseline. \textbf{Top row (TRACE-8):} (a)~OMA and (b)~BWT for pool sizes $P \in \{4, 6, 8\}$. \textbf{Bottom row (CITB-19):} (c)~OMA and (d)~BWT for pool sizes $P \in \{3, 10, 19\}$. Knowledge accumulation (OMA) is successful across all configurations; performance is non-monotonic with respect to pool size on TRACE, while CITB shows comparable cumulative scores but diverging forgetting profiles.}
\label{fig:online_pool_metrics}
\end{figure}

\FloatBarrier
\subsection{Extended Batched Performance}

\subsubsection{Per-Task Batched Performance}
\label{app:pertask_batched}

\begin{figure}[htbp]
    \centering
    \begin{subfigure}[b]{0.46\textwidth}
        \centering
        \includegraphics[width=\textwidth]{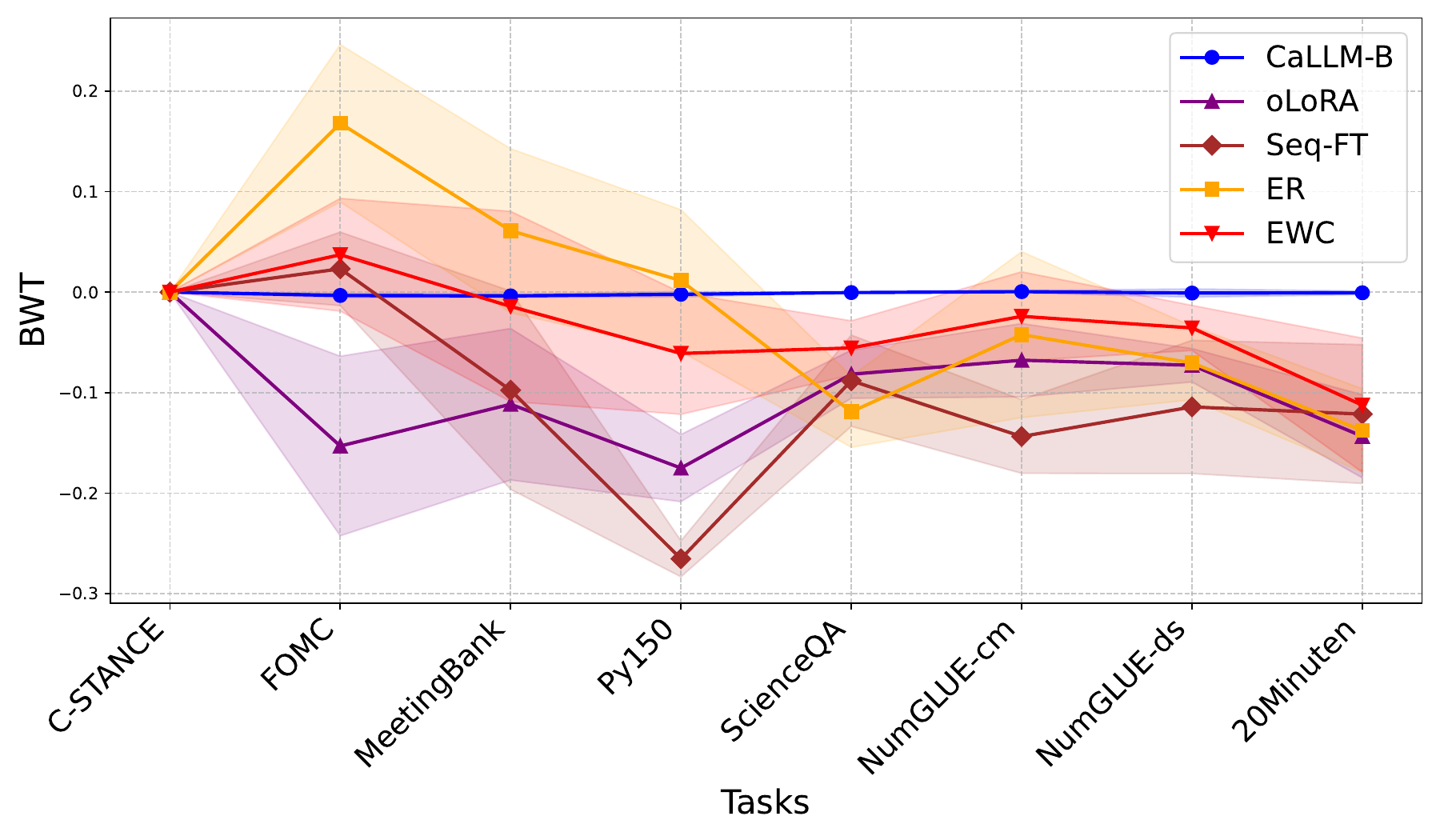}
        \caption{TRACE-8 dataset}
        \label{fig:bwt_trace}
    \end{subfigure}
    \begin{subfigure}[b]{0.53\textwidth}
        \centering
        \includegraphics[width=\textwidth]{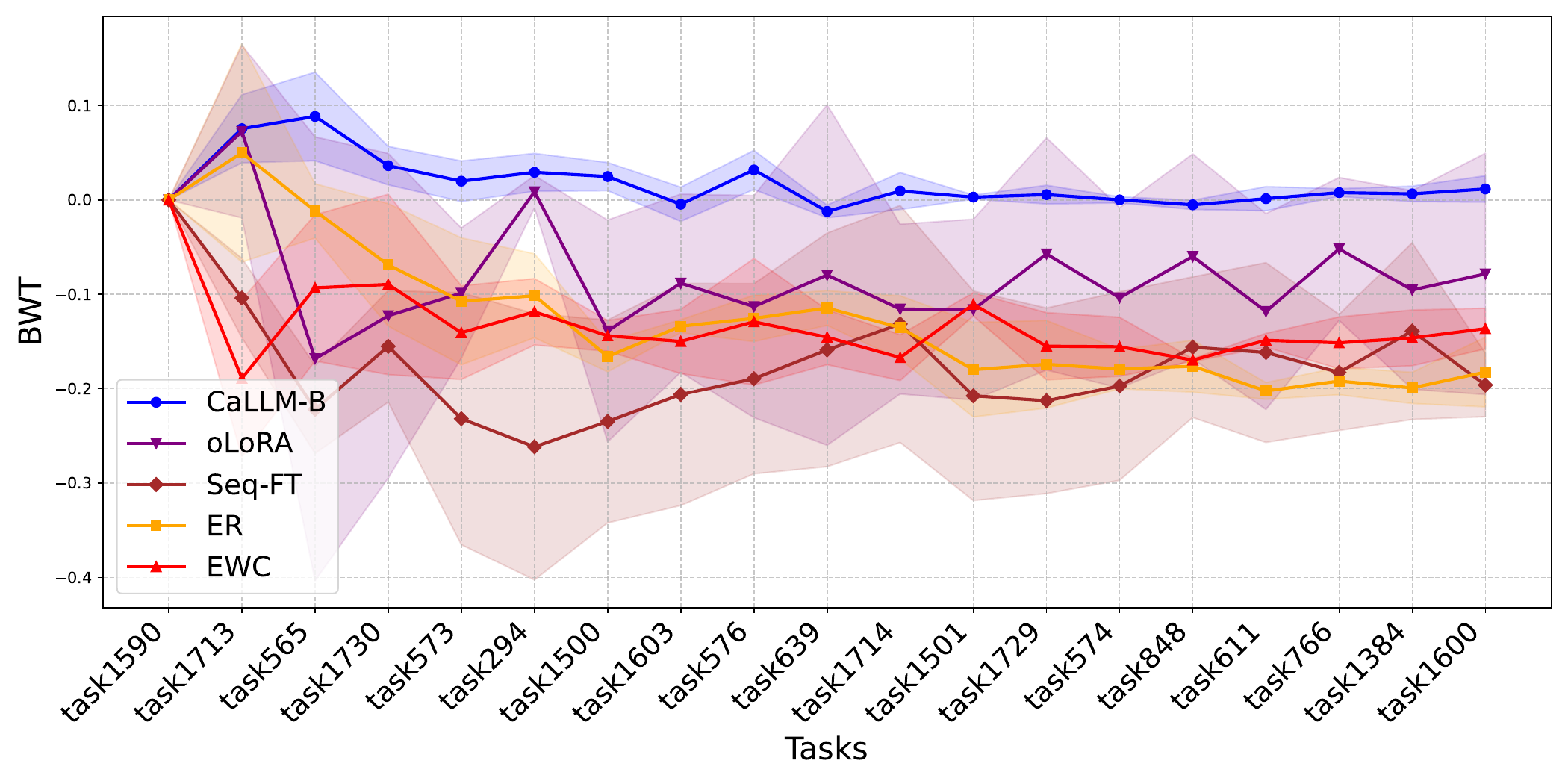}
        \caption{CITB-19 dataset}
        \label{fig:bwt_citb}
    \end{subfigure}
    
    \caption{\textbf{Task-Incremental BWT Comparison Across Baselines (mean $\pm$ std).} CaLLM-B maintains a BWT score close to zero throughout the stream on both benchmarks.}
    \label{fig:bwt_comparison}
\end{figure}

Table~\ref{tab:pertask_trace_batched} and Table~\ref{tab:pertask_citb_batched} show the per-task performance on TRACE-8 and CITB-19 for the CaLLM-B baseline as each task is sequentially presented to the model, capturing the immediate performance right after all training data for that specific task has been processed. These tables are complementary to the results illustrated in Figure~\ref{fig:batched_baseline_comparison}, demonstrating that performance on individual tasks is steadily maintained as the stream length increases. This stability is further supported by Figure~\ref{fig:bwt_comparison}, which shows that CaLLM consistently achieves a BWT score of nearly zero, confirming minimal catastrophic forgetting across the streaming timeline.

\begin{table}[htbp]
\centering
\caption{\textbf{Per-Task Performance Table on TRACE-8 for CaLLM-B}. Results are reported for a single random seed. Metrics used are accuracy (C-STANCE, FOMC, ScienceQA, NumGLUE-cm/ds), ROUGE-L (MeetingBank), EDIM (Py150), and SARI (20Minuten).} 
\label{tab:pertask_trace_batched}
\small
\setlength{\tabcolsep}{4pt}
\begin{tabular}{lcccccccc}
\toprule
\textbf{Task}  & {C-STANCE}& {FOMC}& {MeetingBank}& {Py150}& {ScienceQA}& {NumGLUE-cm}& {NumGLUE-ds}&{20Minuten} \\ 
\midrule
C-STANCE  & 0.42 & 0.42 & 0.42 & 0.42 & 0.41 & 0.42 & 0.42 & 0.42 \\
FOMC  & -- & 0.56 & 0.55 & 0.56 & 0.56 & 0.56 & 0.56 & 0.56 \\
MeetingBank  & -- & -- & 0.22 & 0.23 & 0.23 & 0.23 & 0.23 & 0.23 \\
Py150  & -- & -- & -- & 16.94 & 16.95 & 16.85 & 17.15 & 17.12 \\
ScienceQA  & -- & -- & -- & -- & 0.56 & 0.57 & 0.56 & 0.56 \\
NumGLUE-cm  & -- & -- & -- & -- & -- & 0.19 & 0.19 & 0.19 \\
NumGLUE-ds  & -- & -- & -- & -- & -- & -- & 0.30 & 0.30 \\
20Minuten  & -- & -- & -- & -- & -- & -- & -- & 39.59 \\ 
\bottomrule
\end{tabular}
\end{table}

\begin{table}[htbp]
\centering
\caption{\textbf{Per-Task Performance Table on CITB-19 for CaLLM-B}. Results are reported for a single random seed. The metric used is ROUGE-L for all tasks.} 
\label{tab:pertask_citb_batched}
\small
\setlength{\tabcolsep}{4pt}
\begin{tabular}{lccccccccccccccccccc}
\toprule
\textbf{Task} & \rot{task1590} & \rot{task1713} & \rot{task565}  & \rot{task1730} & \rot{task573}  & 
\rot{task294}  & \rot{task1500} & \rot{task1603} & \rot{task576}  & \rot{task639}  & 
\rot{task1714} & \rot{task1501} & \rot{task1729} & \rot{task574}  & \rot{task848}  & 
\rot{task611}  & \rot{task766}  & \rot{task1384} & \rot{task1600} \\ 
\midrule
task1590 & 0.41 & 0.50 & 0.55 & 0.44 & 0.53 & 0.45 & 0.52 & 0.44 & 0.46 & 0.46 & 0.40 & 0.43 & 0.47 & 0.38 & 0.56 & 0.55 & 0.44 & 0.46 & 0.53 \\
task1713 & --   & 0.60 & 0.52 & 0.56 & 0.53 & 0.59 & 0.59 & 0.51 & 0.56 & 0.53 & 0.55 & 0.64 & 0.52 & 0.53 & 0.52 & 0.49 & 0.56 & 0.49 & 0.54 \\
task565  & --   & --   & 0.37 & 0.41 & 0.38 & 0.39 & 0.33 & 0.31 & 0.37 & 0.33 & 0.29 & 0.29 & 0.21 & 0.36 & 0.37 & 0.29 & 0.45 & 0.29 & 0.33 \\
task1730 & --   & --   & --   & 0.24 & 0.15 & 0.21 & 0.18 & 0.14 & 0.18 & 0.18 & 0.20 & 0.21 & 0.18 & 0.21 & 0.19 & 0.13 & 0.20 & 0.14 & 0.18 \\
task573  & --   & --   & --   & --   & 0.45 & 0.43 & 0.47 & 0.49 & 0.48 & 0.44 & 0.55 & 0.56 & 0.45 & 0.44 & 0.34 & 0.50 & 0.56 & 0.60 & 0.51 \\
task294  & --   & --   & --   & --   & --   & 0.11 & 0.11 & 0.12 & 0.09 & 0.10 & 0.12 & 0.13 & 0.15 & 0.13 & 0.13 & 0.11 & 0.16 & 0.09 & 0.14 \\
task1500 & --   & --   & --   & --   & --   & --   & 0.09 & 0.07 & 0.13 & 0.10 & 0.13 & 0.06 & 0.14 & 0.09 & 0.16 & 0.05 & 0.13 & 0.11 & 0.14 \\
task1603 & --   & --   & --   & --   & --   & --   & --   & 0.70 & 0.80 & 0.76 & 0.62 & 0.70 & 0.74 & 0.76 & 0.61 & 0.66 & 0.65 & 0.59 & 0.67 \\
task576  & --   & --   & --   & --   & --   & --   & --   & --   & 0.08 & 0.10 & 0.09 & 0.07 & 0.07 & 0.10 & 0.07 & 0.07 & 0.07 & 0.07 & 0.08 \\
task639  & --   & --   & --   & --   & --   & --   & --   & --   & --   & 0.25 & 0.29 & 0.20 & 0.25 & 0.22 & 0.23 & 0.24 & 0.18 & 0.23 & 0.23 \\
task1714 & --   & --   & --   & --   & --   & --   & --   & --   & --   & --   & 0.09 & 0.10 & 0.06 & 0.04 & 0.07 & 0.05 & 0.09 & 0.08 & 0.05 \\
task1501 & --   & --   & --   & --   & --   & --   & --   & --   & --   & --   & --   & 0.72 & 0.80 & 0.84 & 0.72 & 0.72 & 0.84 & 0.92 & 0.76 \\
task1729 & --   & --   & --   & --   & --   & --   & --   & --   & --   & --   & --   & --   & 0.08 & 0.08 & 0.20 & 0.15 & 0.08 & 0.12 & 0.04 \\
task574  & --   & --   & --   & --   & --   & --   & --   & --   & --   & --   & --   & --   & --   & 0.09 & 0.04 & 0.08 & 0.08 & 0.09 & 0.09 \\
task848  & --   & --   & --   & --   & --   & --   & --   & --   & --   & --   & --   & --   & --   & --   & 0.27 & 0.27 & 0.23 & 0.28 & 0.29 \\
task611  & --   & --   & --   & --   & --   & --   & --   & --   & --   & --   & --   & --   & --   & --   & --   & 0.31 & 0.28 & 0.23 & 0.21 \\
task766  & --   & --   & --   & --   & --   & --   & --   & --   & --   & --   & --   & --   & --   & --   & --   & --   & 0.18 & 0.21 & 0.25 \\
task1384 & --   & --   & --   & --   & --   & --   & --   & --   & --   & --   & --   & --   & --   & --   & --   & --   & --   & 0.04 & 0.04 \\
task1600 & --   & --   & --   & --   & --   & --   & --   & --   & --   & --   & --   & --   & --   & --   & --   & --   & --   & --   & 0.30 \\ 
\bottomrule
\end{tabular}
\end{table}

\subsubsection{CITB Extended Batched Performance}
\label{app:citb_per_task}

Table~\ref{tab:citb_per_task} disaggregates the final CITB-19 results by individual task, analogous to Table~\ref{tab:trace_per_task} for TRACE-8. All 19 tasks use ROUGE-L (0--1) as the evaluation metric. Scores are averaged across 3 seeds for all methods except EWC (2 seeds available). CaLLM-B achieves the highest score among all continual learning methods on 13 of 19 tasks. On the remaining 6, oLoRA leads on individual tasks but trails CaLLM-B in mean performance (0.24 vs.\ 0.31), indicating that its per-task wins come at the cost of broader forgetting elsewhere. The high Inference-Only scores on tasks such as Diplomacy (T1590), P.Chat Gen.\ (T1729), and Mutual Dial.\ (T611) reflect strong backbone priors that sequential fine-tuning erodes for all methods.

\begin{table}[h]
  \centering
  \caption{\textbf{Per-Task Final Performance on CITB-19 (Batched).}
  Scores measured after all 19 tasks have been trained sequentially; all tasks use ROUGE-L (0--1) averaged over 3 seeds.}
  \label{tab:citb_per_task}
  \small
  \resizebox{\linewidth}{!}{%
  \begin{tabular}{lcccccc}
    \toprule
    \textbf{Task} & \textbf{Inf-Only} & \textbf{Seq-FT} & \textbf{EWC} & \textbf{ER} & \textbf{oLoRA} & \textbf{CaLLM-B} \\
    \midrule
    Diplomacy (T1590)         & \textbf{0.84} & 0.14 & 0.05 & 0.20 & 0.10 & 0.31 \\
    ConvAI3-1713              & \textbf{0.36} & 0.05 & 0.07 & 0.14 & 0.18 & 0.28 \\
    CIRCA (T565)              & 0.08 & 0.10 & 0.04 & 0.09 & 0.28 & \textbf{0.40} \\
    P.Chat Choose (T1730)     & 0.02 & 0.24 & 0.02 & 0.09 & 0.39 & \textbf{0.45} \\
    AirDial.\ Class.\ (T573)  & \textbf{0.92} & 0.05 & 0.07 & 0.08 & 0.27 & 0.36 \\
    Story Motiv.\ (T294)      & 0.34 & 0.09 & 0.07 & 0.09 & \textbf{0.49} & 0.25 \\
    DSTC3 Class.\ (T1500)     & 0.12 & 0.05 & 0.00 & 0.03 & 0.12 & \textbf{0.44} \\
    SMCalFlow-1603            & \textbf{0.68} & 0.12 & 0.06 & 0.12 & 0.11 & 0.47 \\
    Curiosity Dial.\ (T576)   & 0.25 & 0.28 & 0.03 & 0.04 & \textbf{0.29} & 0.28 \\
    MultiWOZ (T639)           & 0.14 & 0.07 & 0.10 & 0.08 & \textbf{0.32} & 0.15 \\
    ConvAI3-1714              & 0.20 & 0.11 & 0.02 & 0.23 & 0.12 & \textbf{0.25} \\
    DSTC3 Ans.\ (T1501)       & 0.06 & 0.05 & 0.03 & 0.07 & 0.31 & \textbf{0.41} \\
    P.Chat Gen.\ (T1729)      & \textbf{0.82} & 0.16 & 0.08 & 0.07 & 0.18 & 0.09 \\
    AirDial.\ Gen.\ (T574)    & \textbf{0.22} & 0.11 & 0.05 & 0.09 & 0.16 & 0.20 \\
    PubMedQA (T848)           & 0.08 & 0.17 & 0.04 & 0.05 & 0.06 & \textbf{0.21} \\
    Mutual Dial.\ (T611)      & \textbf{0.84} & 0.05 & 0.12 & 0.19 & 0.36 & 0.26 \\
    Craigslist (T766)         & 0.21 & 0.13 & 0.05 & 0.12 & 0.11 & \textbf{0.36} \\
    DealOrNo (T1384)          & 0.07 & 0.12 & 0.10 & 0.10 & 0.26 & \textbf{0.38} \\
    SMCalFlow-1600            & 0.26 & 0.36 & 0.19 & 0.28 & \textbf{0.50} & 0.35 \\
    \bottomrule
  \end{tabular}%
  }
\end{table}

\FloatBarrier
\subsection{Inference-Only baseline: prompt-template sensitivity}
\label{app:prompt-sensitivity}
\begin{figure}
    \centering
    \includegraphics[width=\linewidth]{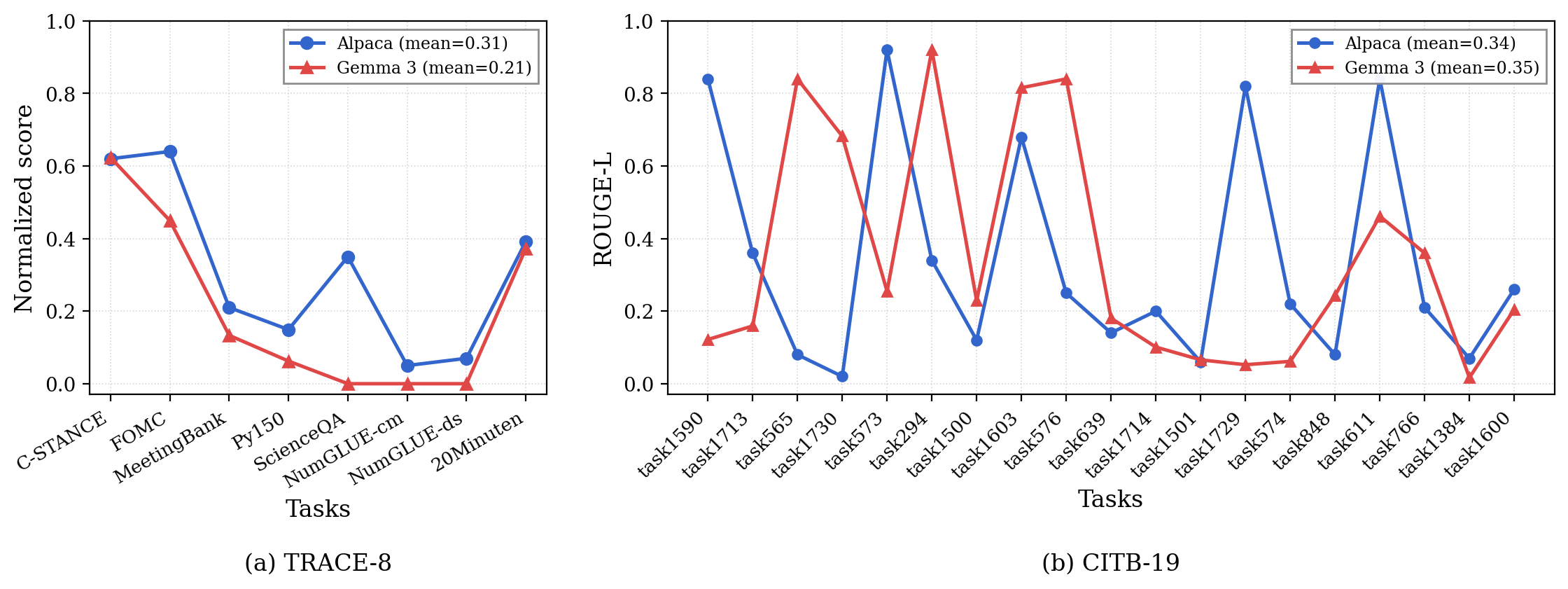}
    \caption{\textbf{Prompt-template sensitivity of the Inference-Only
    baseline.} Per-task final performance under the Alpaca template (blue,
    used in the main paper) and Gemma~3's native chat template (red).
    \textbf{(a) TRACE-8 (Batched):} normalized scores (EDIM, SARI divided by
    100); Alpaca dominates on every task (mean $0.31$ vs.\ $0.21$).
    \textbf{(b) CITB-19 (Batched):} ROUGE-L; the two templates are tied in
    aggregate (mean $0.34$ vs.\ $0.35$) but redistribute mass across
    generation- and classification-style tasks. Tasks are shown in original
    training/streaming order.}
    \label{fig:prompt-sensitivity}
\end{figure}

To test the robustness of the performance to the prompt template, we repeated the Inference-Only baseline on both TRACE-8 and CITB-19 using Gemma 3's native chat format (\texttt{<start\_of\_turn>user~\ldots<end\_of\_turn><start\_of\_turn>model}), keeping every other hyperparameter identical to the Alpaca-format run reported in the main paper. We present the per-task comparison in Figure~\ref{fig:prompt-sensitivity}. On TRACE-8, the Alpaca template is strictly stronger or equal on every task (mean normalized score $0.31$ vs.\ $0.21$), with the largest drops on structured-output tasks (Py150, ScienceQA, NumGLUE-cm/ds). On CITB-19 the two templates are tied in aggregate (mean ROUGE-L $0.34$ vs.\ $0.35$) but redistribute mass between generation-style tasks (which lose $0.4$--$0.8$), and classification-/choice-style tasks, which gain a comparable amount. We therefore retain the Alpaca-format Inf-Only as the headline baseline as it is the more conservative comparison on TRACE-8 and equivalent on CITB-19.

\FloatBarrier
\section{Extended Router Analysis}

This section provides a deep dive into the router's internal logic, visualizing how it represents tasks and how capacity constraints ($P$) dictate expert assignment.

\FloatBarrier
\subsection{Task Separability in the Embedding Space} 
\begin{figure}[h]
\begin{center}
\includegraphics[width=0.45\linewidth]{figs/trace_router_embeddings.png}
\includegraphics[width=0.45\linewidth]{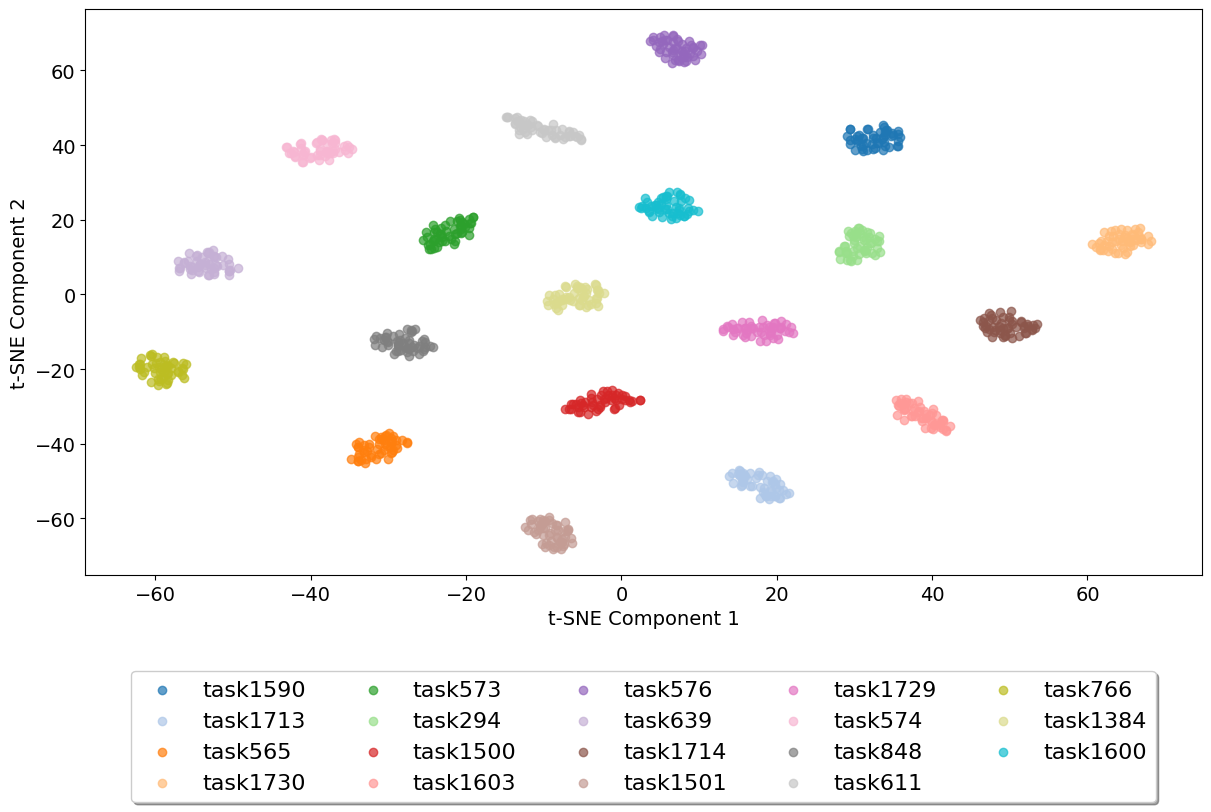}
\end{center}
\caption{\textbf{Router Embedding Space (2D Projection).} t-SNE projections of the router's embeddings, coloured by task identity. \textbf{(a)}~TRACE-8: 8 tasks spanning diverse domains. \textbf{(b)}~CITB-19: 19 dialogue sub-tasks.}
\label{fig:router_embeddings_complete}
\end{figure}

Figure~\ref{fig:router_embeddings_complete} illustrates the inherent separability of the task distributions within the router's embedding space. Using t-SNE projections of the frozen backbone's mean-pooled features, we observe that diverse tasks (both for TRACE-8 and CITB-19) form well-defined, distant clusters.

\FloatBarrier
\subsection{Prototype Dynamics and Capacity Progression} \label{app:proto_embeddings}
\begin{figure}[h]
  \centering
  \begin{subfigure}[b]{0.48\linewidth}
    \centering
    \includegraphics[width=\linewidth]{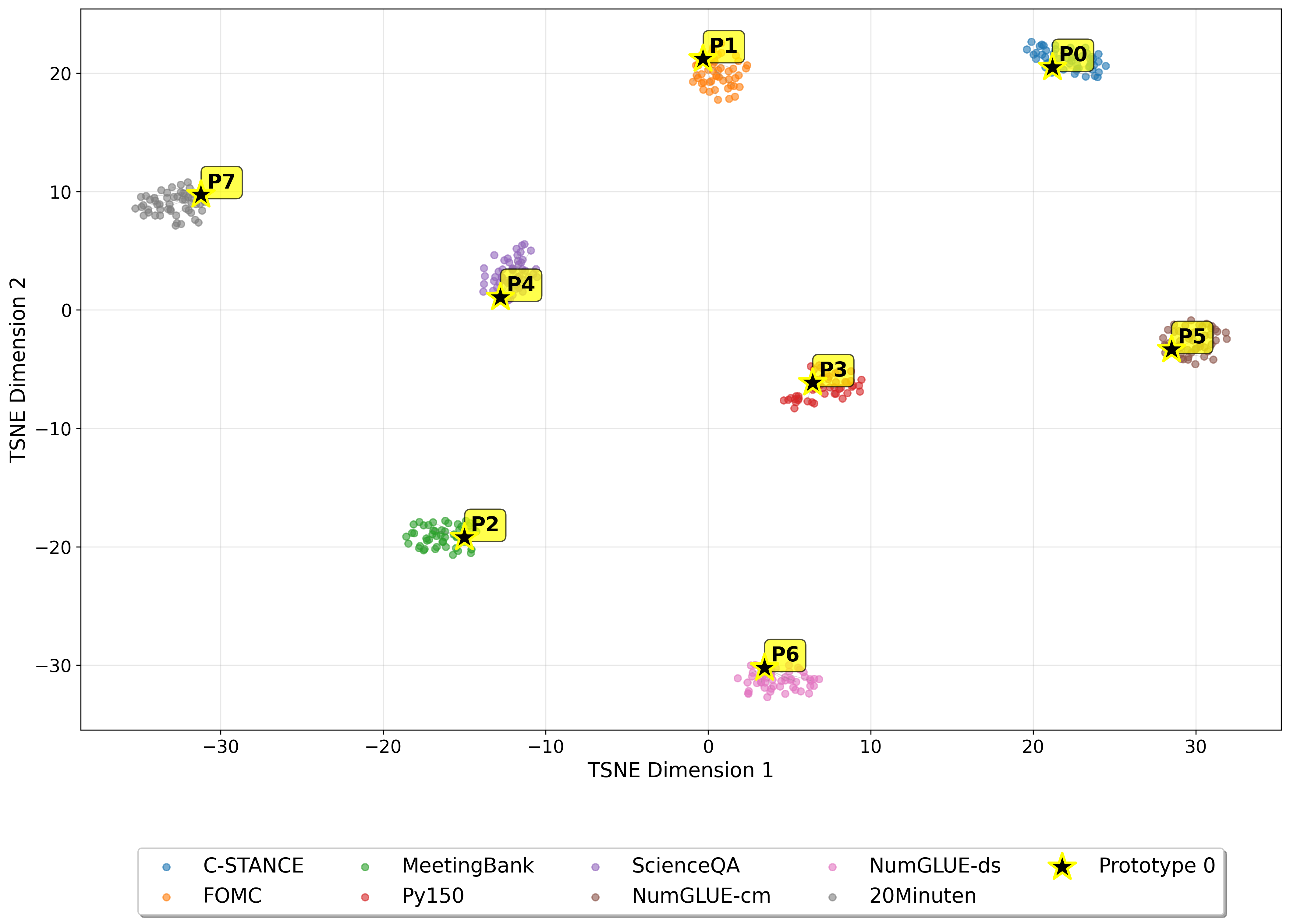}
    \caption{$P{=}8$: perfect isolation, one prototype per task cluster.}
    \label{fig:proto_embed_8}
  \end{subfigure}\hfill
  \begin{subfigure}[b]{0.48\linewidth}
    \centering
    \includegraphics[width=\linewidth]{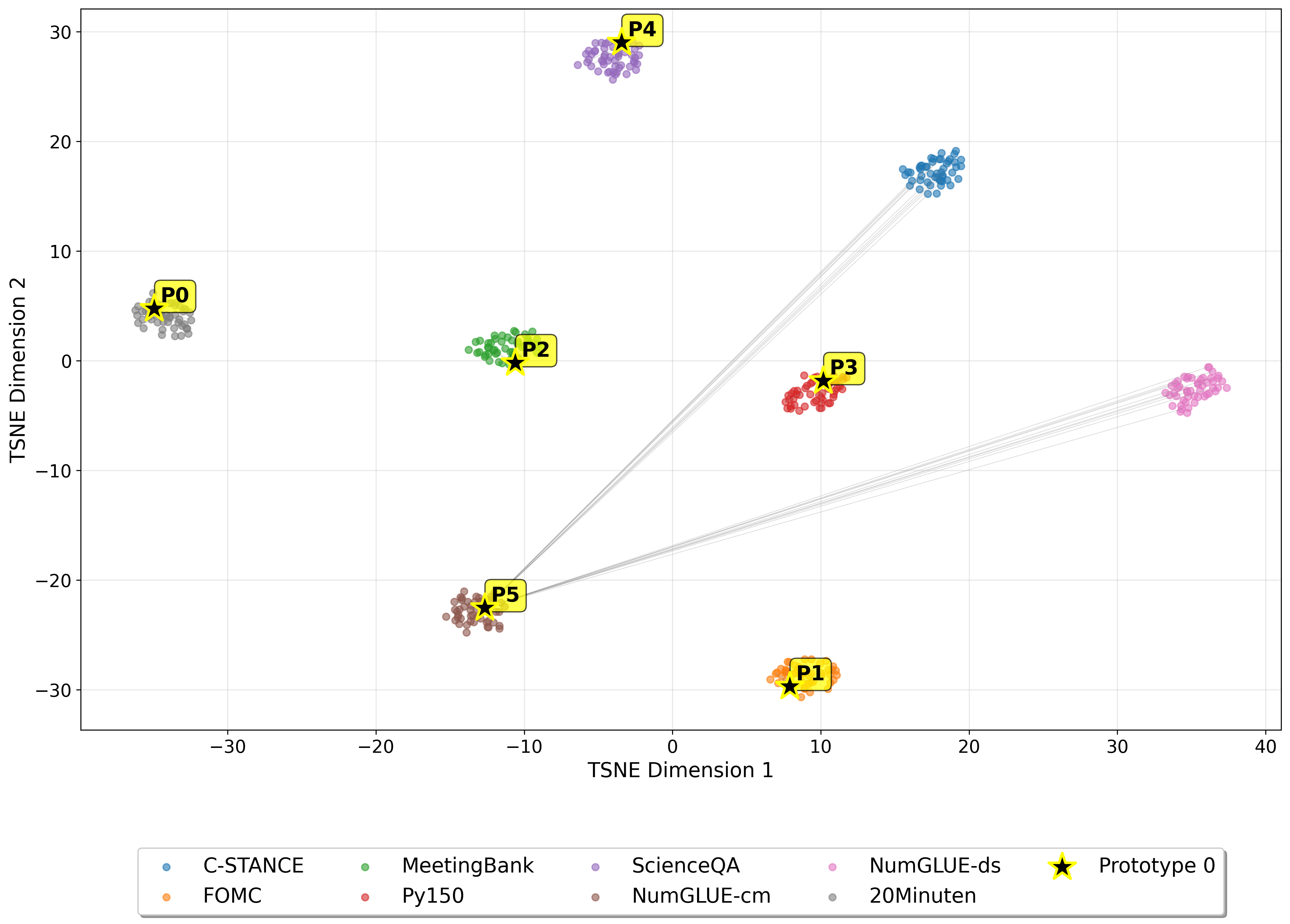}
    \caption{$P{=}6$: NumGLUE-cm/ds share a prototype---constructive merge, negligible BWT.}
    \label{fig:proto_embed_6}
  \end{subfigure}

  \vspace{0.5em}

  \begin{subfigure}[b]{0.48\linewidth}
    \centering
    \includegraphics[width=\linewidth]{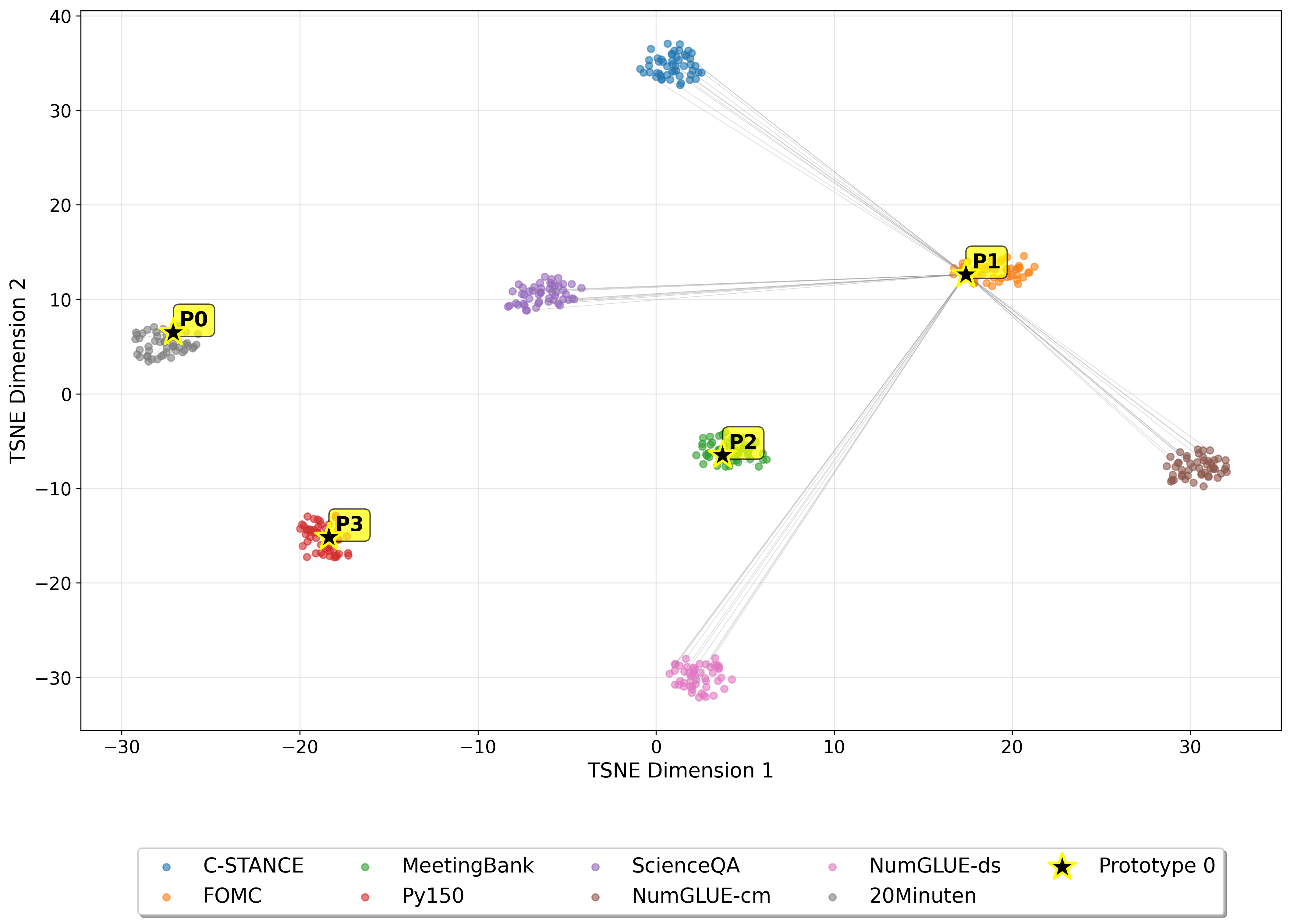}
    \caption{$P{=}4$: one prototype absorbs 5 diverse tasks---destructive overload, BWT $= -0.19$.}
    \label{fig:proto_embed_4}
  \end{subfigure}\hfill
  \begin{subfigure}[b]{0.48\linewidth}
    \centering
    \includegraphics[width=\linewidth]{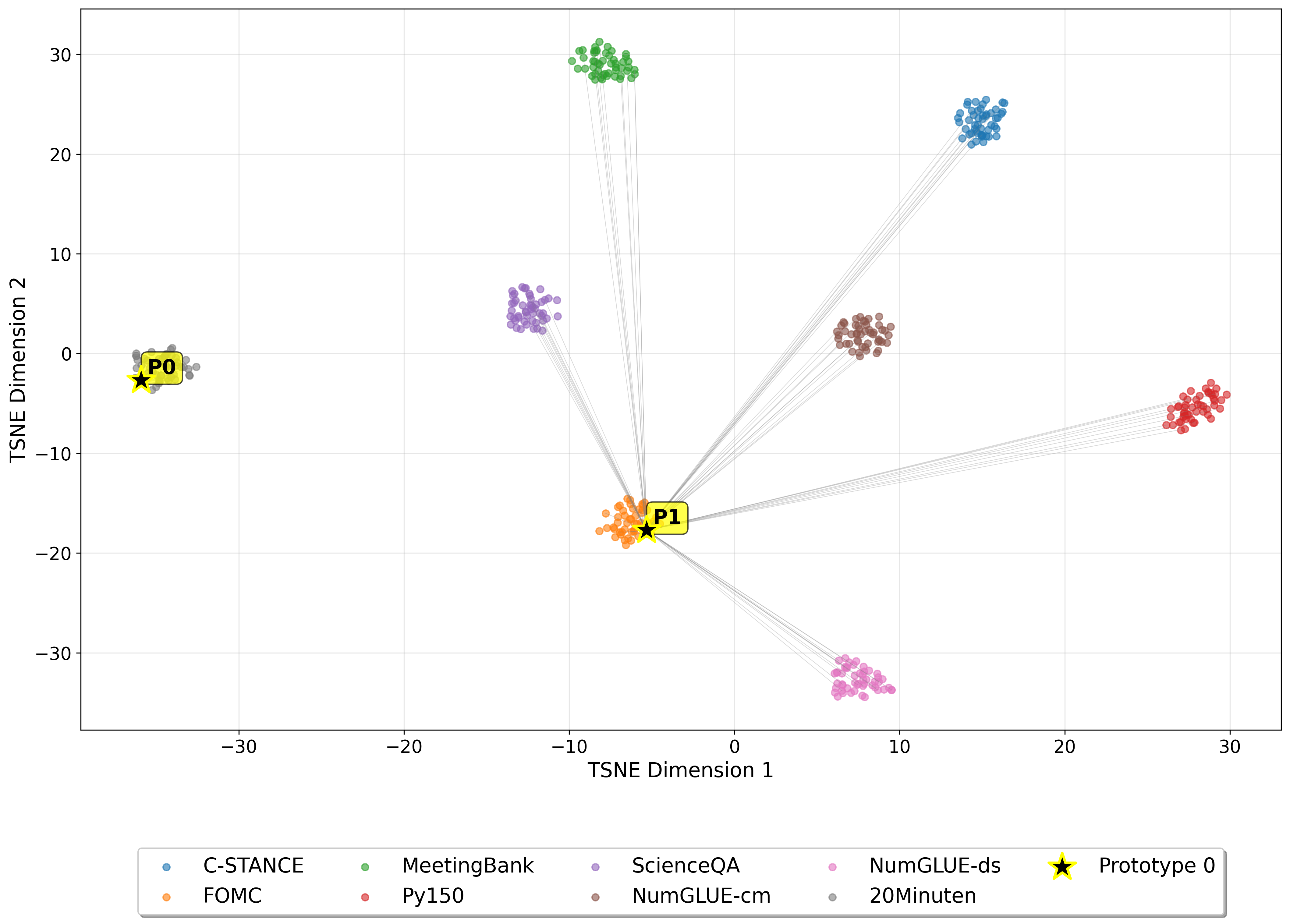}
    \caption{$P{=}2$: system collapse, 7 of 8 tasks share one prototype, BWT $= -0.25$.}
    \label{fig:proto_embed_2}
  \end{subfigure}
  \caption{\textbf{Router Embedding Space with Prototypes (TRACE-8, t-SNE).} Yellow stars are prototype positions; grey lines show nearest-prototype assignment. The transition from $P{=}8$ to $P{=}2$ maps directly onto the isolation $\to$ constructive clustering $\to$ destructive overload $\to$ collapse progression described in \S\ref{sec:capacity_dynamics}.}
  \label{fig:router_embeddings_proto}
\end{figure}

Figure~\ref{fig:router_embeddings_proto} demonstrate how the router dynamically partitions the embedding space under varying capacity constrains, studied in \S\ref{sec:capacity_dynamics}, for TRACE-8 . Each panel overlays $P$ prototypes (yellow stars) on the task-cluster structure; grey lines connect each data point to its nearest prototype. The prototype positions determine which tasks share an expert during training.

At $P{=}8$ (Figure~\ref{fig:proto_embed_8}), every prototype settles at the centroid of one task cluster, giving zero-interference isolation. At $P{=}6$ (Figure~\ref{fig:proto_embed_6}), the two NumGLUE math tasks and the C-STANCE dataset are pulled toward a shared prototype, a constructive merge that incurs negligible BWT. At $P{=}4$ (Figure~\ref{fig:proto_embed_4}), one prototype absorbs five semantically diverse tasks on the right side of the space, producing the conflicting gradients that drive destructive overload. Even here, Py150 (code) retains its own isolated prototype (in red), confirming the router's sensitivity to the most extreme distributional gap. At $P{=}2$ (Figure~\ref{fig:proto_embed_2}), only 20Minuten keeps a dedicated prototype; all other tasks collapse onto a single expert.

\FloatBarrier
\subsection{Online Streaming Heatmap}
\label{app:online_heatmap}
\begin{figure}[h]
  \centering
  \includegraphics[width=\linewidth]{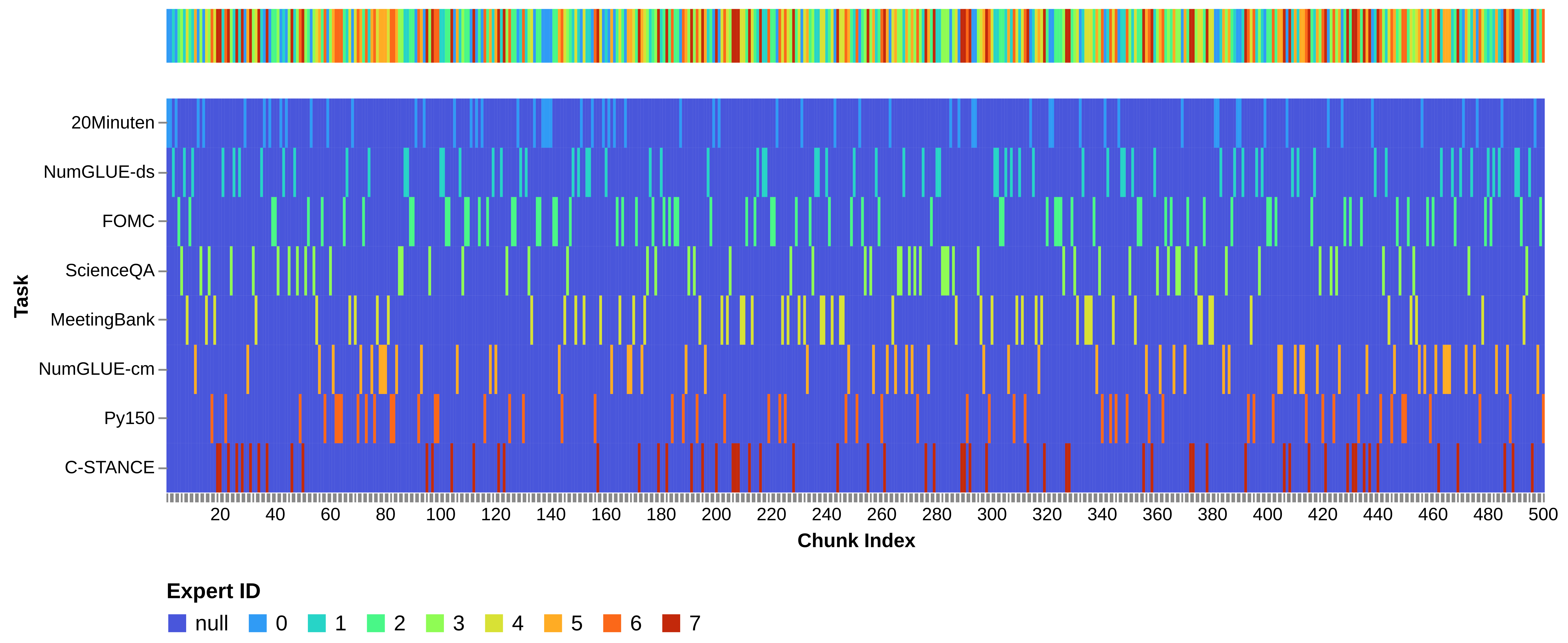}
  \caption{\textbf{Full Expert-Assignment Heatmap (TRACE, 500 chunks, P=8).} Each row is a task; each column a chunk; colour encodes the assigned expert ID. Dark blue (``null'') cells indicate chunks where that task was absent from the stream. The per-task colour patterns remain stable across the full 500-chunk stream, confirming persistent task--expert binding.}
  \label{fig:online_heatmap_full}
\end{figure}

While the previous plots show the router assignment in the Batched scenario,  Figure~\ref{fig:online_heatmap_full} shows the complete 500-chunk task$\times$expert assignment heatmap for the TRACE online stream. Each row corresponds to a task, each column to a chunk, and colour encodes the expert ID assigned by the router. The ``null'' entries (dark blue) indicate chunks in which the corresponding task did not appear in the interleaved stream. Despite the sparsity, the per-task colour patterns are stable throughout the full stream, confirming the persistent task---expert binding reported in §\ref{sec:online_results}. The zoomed view (chunks~200--300) used in the main text (Figure~\ref{fig:online_trace}(b)) was selected as a representative window that minimises null-cell density and maximises readability.

\FloatBarrier
\subsubsection{Aggregate Online Routing Matrices} \label{app:confusion_matrices}
\begin{figure}[h]
  \centering
  \includegraphics[width=\linewidth]{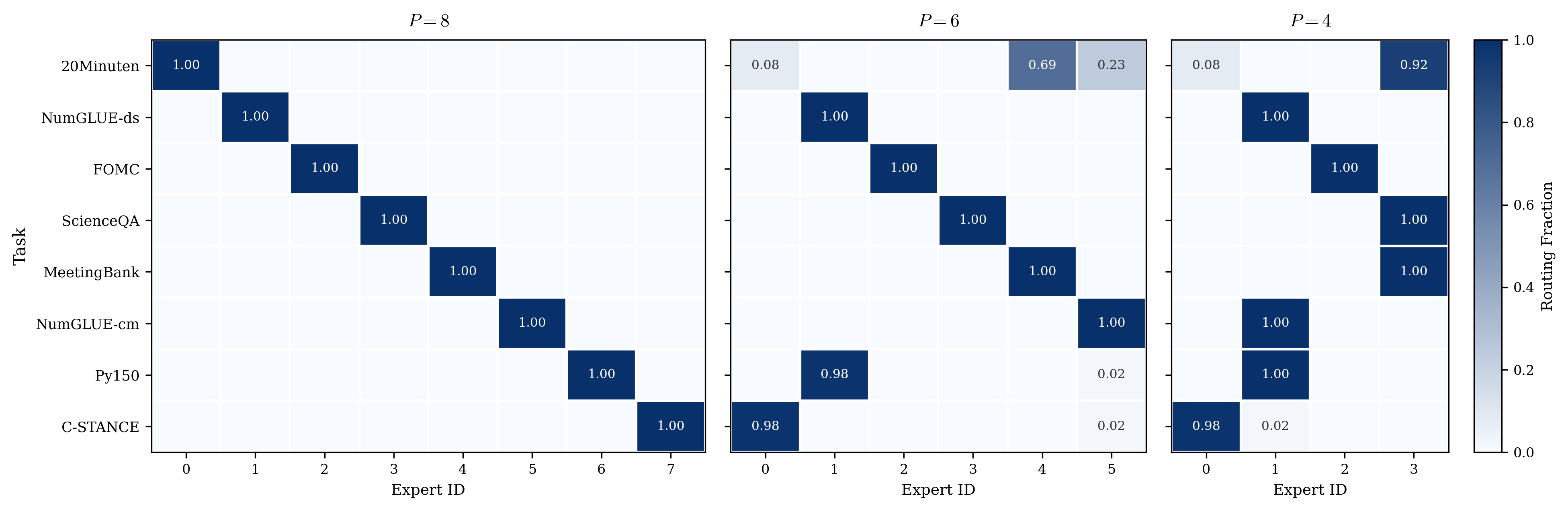}
  \caption{\textbf{Router Assignment Matrices on TRACE (Online).} Pool sizes $P \in \{4,6,8\}$. Each cell shows the fraction of chunks from that task routed to each expert.}
  \label{fig:confusion_matrices}
\end{figure}

The temporal stability observed in the full stream is summarized numerically in Figure~\ref{fig:confusion_matrices}. These matrices for TRACE ($P \in \{4, 6, 8\}$) show the fraction of chunks routed to each expert over a 500-chunk stream. As analyzed in \S\ref{sec:online_pool}, these matrices reveal that at $P{=}4$, the router settles into highly stable, semantically aligned groups that outperform the $P{=}6$ configuration due to reduced task displacement.

\end{document}